\documentclass[lettersize,journal]{IEEEtran}
\usepackage{amsmath,amsfonts}
\usepackage{algorithmic}
\usepackage{algorithm}
\usepackage{array}
\usepackage{textcomp}
\usepackage{stfloats}
\usepackage{url}
\usepackage{verbatim}
\usepackage{graphicx}
\usepackage{cite}
\usepackage{color}
\usepackage{microtype}
\usepackage{graphicx}
\usepackage{booktabs}
\usepackage{hyperref}
\usepackage{tcolorbox}
\usepackage{amssymb}
\usepackage{mathtools}
\usepackage{amsthm}
\usepackage{caption}
\usepackage{pifont}
\usepackage{subcaption}
\usepackage[capitalize,noabbrev]{cleveref}

\theoremstyle{plain}
\newtheorem{theorem}{Theorem}

\theoremstyle{definition}

\theoremstyle{remark}

\newtheorem{mydef}{\bf{Definition}}

\DeclareMathAlphabet\mathbfcal{OMS}{cmsy}{b}{n}

\usepackage[textsize=tiny]{todonotes}

\begin{document}

\title{Bridging Models to Defend: A Population-Based Strategy for Robust Adversarial Defense}

\author{Ren~Wang,~\IEEEmembership{Member,~IEEE,}  Yuxuan~Li,~\IEEEmembership{Student Member,~IEEE,}
Can~Chen,~\IEEEmembership{Member,~IEEE,}        Dakuo~Wang,~\IEEEmembership{Senior~Member,~IEEE,} Jinjun~Xiong,~\IEEEmembership{Fellow,~IEEE,} Pin-Yu~Chen,~\IEEEmembership{Fellow,~IEEE,}        Sijia~Liu,~\IEEEmembership{Senior~Member,~IEEE,} Mohammad~Shahidehpour,~\IEEEmembership{Life~Fellow,~IEEE,}       and~Alfred~Hero,~\IEEEmembership{Life~Fellow,~IEEE}
\IEEEcompsocitemizethanks{\IEEEcompsocthanksitem Ren Wang is with the Department
of Electrical and Computer Engineering, Illinois Institute of Technology, Chicago,
IL 60616.
\IEEEcompsocthanksitem Yuxuan Li is a graduate research intern in the Department
of Electrical and Computer Engineering, Illinois Institute of Technology, Chicago,
IL 60616.
\IEEEcompsocthanksitem Can Chen is with the School of Data Science and Society, University of North Carolina at Chapel Hill, Chapel Hill, NC 27599.
\IEEEcompsocthanksitem Dakuo Wang is with the Khoury College of Computer Sciences and the College of Arts, Media and Design, Northeastern University, Boston, MA 02115.
\IEEEcompsocthanksitem Jinjun Xiong is with the Department of Computer Science and Engineering, University at Buffalo,
Buffalo, NY 14260.
\IEEEcompsocthanksitem Pin-Yu Chen is with the IBM Thomas J. Watson Research Center, NY 10598.
\IEEEcompsocthanksitem Sijia Liu is with the Department of Computer Science and Engineering,
Michigan State University, East Lansing, MI 48824.
\IEEEcompsocthanksitem Mohammad Shahidehpour is with the Department
of Electrical and Computer Engineering, Illinois Institute of Technology, Chicago,
IL 60616.
\IEEEcompsocthanksitem Alfred Hero is with the Electrical Engineering and Computer Science Department,
University of Michigan, Ann Arbor, MI 48109.}
\thanks{The first two authors contributed equally to this paper.}
\thanks{Corresponding author: Ren Wang. E-mail: rwang74@iit.edu}
\thanks{Early versions of this work partially appeared in the conference proceedings  \cite{wang2023exploring} and \cite{wang2024deep}. 
}
\thanks{This work was supported in part by the National Science Foundation under grants CCF-2450414,  IIS-2246157, FMitF-2319243, by the Department of Energy under grant DE-CR0000042, and by the US Army Research Office under grant W911NF2310343.}
}

\markboth{Journal of \LaTeX\ Class Files,~Vol.~14, No.~8, August~2021}%
{Shell \MakeLowercase{\textit{et al.}}: A Sample Article Using IEEEtran.cls for IEEE Journals}


\maketitle

\begin{abstract}
Adversarial robustness is a critical measure of a neural network's ability to withstand adversarial attacks at inference time. While robust training techniques have improved defenses against individual $\ell_p$-norm attacks (e.g., $\ell_2$ or $\ell_\infty$), models remain vulnerable to diversified $\ell_p$ perturbations. To address this challenge, we propose a novel Robust Mode Connectivity (RMC)-oriented adversarial defense framework comprising two population-based learning phases. In Phase I, RMC searches the parameter space between two pre-trained models to construct a continuous path containing models with high robustness against multiple $\ell_p$ attacks. To improve efficiency, we introduce a Self-Robust Mode Connectivity (SRMC) module that accelerates endpoint generation in RMC. Building on RMC, Phase II presents RMC-based optimization, where RMC modules are composed to further enhance diversified robustness. To increase Phase II efficiency, we propose Efficient Robust Mode Connectivity (ERMC), which leverages $\ell_1$- and $\ell_\infty$-adversarially trained models to achieve robustness across a broad range of $p$-norms. An ensemble strategy is employed to further boost ERMC’s performance. Extensive experiments across diverse datasets and architectures demonstrate that our methods significantly improve robustness against $\ell_\infty$, $\ell_2$, $\ell_1$, and hybrid attacks. Code is available at \url{https://github.com/wangren09/MCGR}.
\end{abstract}


\section{Introduction}\label{sec:introduction}
The past decade has witnessed rapid advances in deep learning, leading to widespread adoption in high-stakes domains such as medical imaging \cite{sarvamangala2022convolutional}, defect detection \cite{jiwei2019bottom}, and power systems \cite{li2023physics}, where security is critical. Neural networks (NNs), the core of modern deep learning, learn complex mappings from data but remain highly sensitive to small, often imperceptible, input perturbations known as adversarial examples \cite{goodfellow2014explaining,wang2022ask}. Although nearly imperceptible to humans, these perturbations can cause severe model failures, raising serious concerns about the trustworthiness of NNs in safety-critical applications \cite{madry2018towards,carlini2017towards}.

\begin{figure}[h]
  \centering
  \includegraphics[trim=0 0 0 0,clip,width=.49\textwidth]{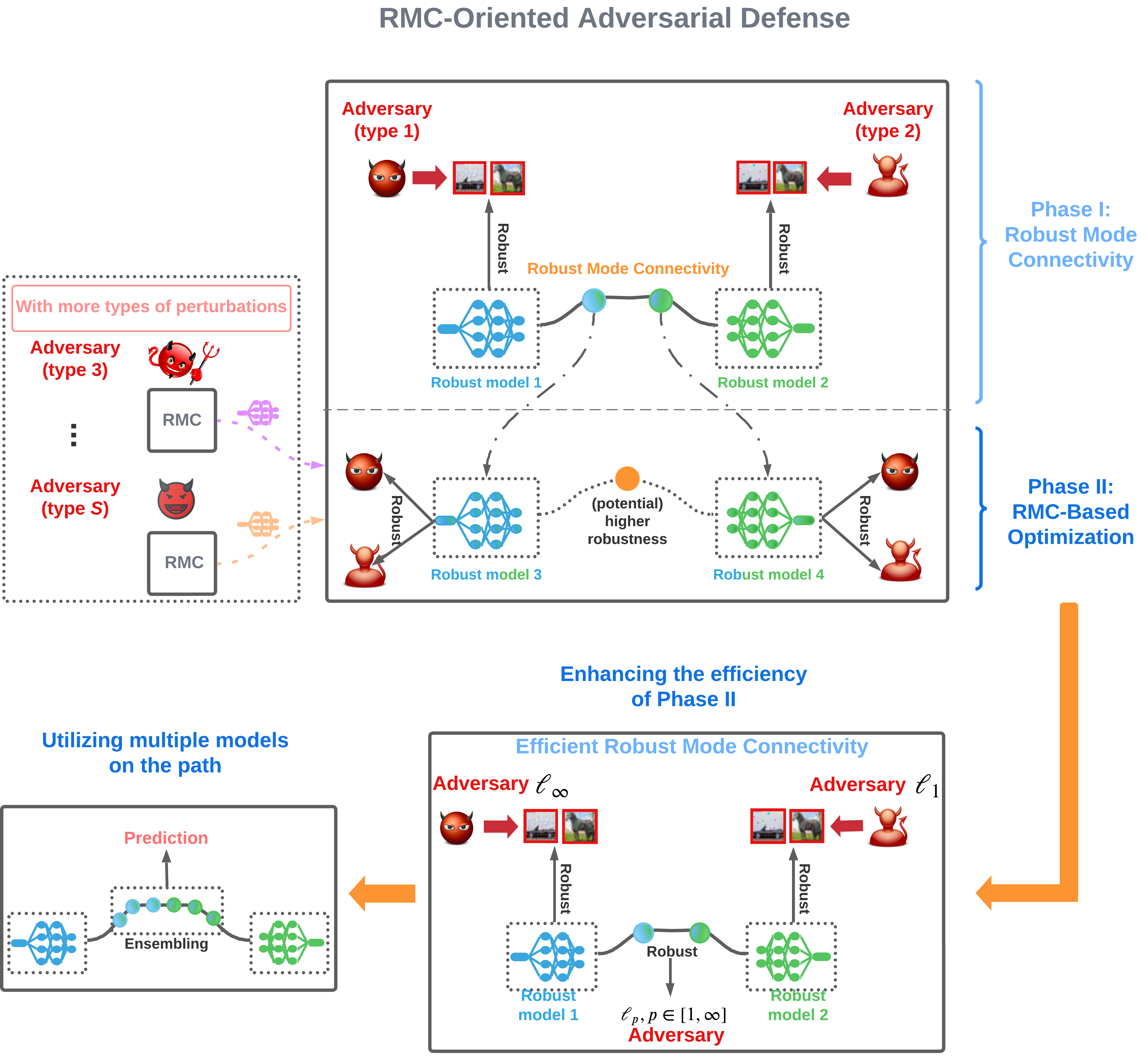}
  \caption{{Overview of the Robust Mode Connectivity (RMC)-Oriented Adversarial Defense Framework}. The upper level of the panel at the top shows Phase I, illustrating that a robust path (robust to adversary types 1 and 2) in the parameter space can be found by connecting one model robust to adversary type 1 and the second model robust to adversary type 2. Selecting optimal points from the path and implementing the RMC process again can further improve robustness, as illustrated in the lower level of the panel at the top. Phase II suggests that more adversary types can be considered by using RMC as the basic unit. The right side panel at the bottom shows the efficient robust mode connectivity (ERMC), which interlaces $\ell_1$ and $\ell_\infty$ robustness into mode connectivity's structure and extends protection to perturbations from $\ell_p \in [1,\infty)$ norms. The left side panel at the bottom illustrates an ensembling method that can further boost the performance of the defense.} 
  \label{fig: framework}
\end{figure}

To address this vulnerability, adversarial training (AT) and its variants have become the most prominent defenses \cite{madry2018towards,zhang2019theoretically,shafahi2019adversarial,wang2020fast}. AT updates model parameters using adversarial examples generated on-the-fly from clean data, enabling the network to learn from adversarial distributions and become more robust during inference. However, most AT methods are designed for a single $\ell_p$ norm constraint (e.g., $\ell_\infty$), and their robustness often degrades sharply under perturbations from other norms \cite{tramer2019adversarial}. While recent works attempt to address this by training on multiple $\ell_p$ norms \cite{croce2019provable,stutz2020confidence,tramer2019adversarial,maini2020adversarial,croce2022adversarial,wang2021adversarial}, they often fall short due to the inherent limitations of single-point optimization in the model parameter space. These approaches can get trapped in local minima or saddle points when optimizing for multiple robustness objectives simultaneously.

In contrast, population-based optimization maintains a diverse set of candidate solutions, enabling broader exploration of the parameter space and better handling of complex, multi-objective tasks such as diversified $\ell_p$ robustness \cite{eiben2015evolutionary,diaz2016review,mirjalili2019evolutionary}. One particularly promising avenue is mode connectivity, which reveals that low-loss, high-accuracy paths often exist between independently trained models \cite{ren2025revisiting,garipov2018loss,freeman2017topology}. This property offers an accelerated population-based strategy for generating many viable models. However, naively applying mode connectivity is insufficient in adversarial settings.

In this work, we aim to improve a model’s robustness against perturbations constrained by different $\ell_p$ norms, with experimental focus on $p = 1, 2, \infty$. Motivated by the limitations of traditional approaches and the promise of mode connectivity, we propose a robust mode connectivity-oriented adversarial defense framework built on population-based optimization.

In Phase I, we introduce Robust Mode Connectivity (RMC), which finds high-robustness paths between adversarially trained models using a multi-steepest descent (MSD) algorithm \cite{maini2020adversarial}. To improve efficiency, we incorporate a Self-Robust Mode Connectivity (SRMC) module, which accelerates the creation of path endpoints. In Phase II, we construct RMC-based optimization, a broader framework that composes RMC modules to generate a population of candidate models and select those with the highest diversified robustness. Further, motivated by theoretical insights that affine classifiers robust to both $\ell_1$ and $\ell_\infty$ attacks can generalize to a wide range of $\ell_p$ threats, we propose Efficient Robust Mode Connectivity (ERMC). This method combines $\ell_1$- and $\ell_\infty$-robust models using a mode connectivity path and ensemble aggregation, boosting efficiency and robustness across norms (Fig.~\ref{fig: framework}).

\noindent\textit{Contributions.}
We summarize our main contributions as follows:

\begin{itemize}
    \item Robust Mode Connectivity (RMC): We propose RMC to construct paths between adversarially trained models, yielding intermediate models with high robustness to diversified $\ell_p$ perturbations. We further introduce Self-Robust Mode Connectivity (SRMC) to accelerate endpoint generation, improving the training efficiency of RMC.
(See Figures~\ref{fig: adv_mc}, \ref{fig: rmc_vgg16_cifar100}, \ref{fig: adv_self_rmc})
    \item RMC-Based Optimization: We extend RMC to a multi-stage population-based optimization framework that further improves robustness by selecting optimal models across multiple RMC units.
(Figures~\ref{fig: adv_mc_opt1}, \ref{fig: adv_mc_opt12}, \ref{fig: adv_mc_opt2})
    \item Efficient Robust Mode Connectivity (ERMC): We propose ERMC, a theoretically grounded method combining $\ell_1$ and $\ell_\infty$ robustness via mode connectivity and ensemble learning, to enhance the efficiency of the RMC-Based Optimization.
(Figure~\ref{fig: adv_self_rmc2})
    \item Comprehensive Evaluation: We conduct extensive experiments demonstrating that RMC, RMC-based optimization, and ERMC significantly outperform existing methods in achieving diversified $\ell_p$ robustness.
(Table~\ref{tab: main})
\end{itemize}

\noindent The rest of this article is organized as follows. Section~\ref{sec: related_work} introduces related works on defenses against diversified $\ell_p$ norm perturbations and population-based neural network learning. In Section~\ref{sec: pre}, we provide the definition of diversified $\ell_p$ robustness, and give introductions to adversarial attack, adversarial training, and mode connectivity. Sections~\ref{sec: rmc} and \ref{sec: opt} introduce the two phases of the proposed mode connectivity-oriented adversarial defense. The RMC method is presented in Section~\ref{sec: rmc}. The RMC-based optimization is proposed in Section~\ref{sec: opt}, and is enhanced by the ERMC method introduced in Section~\ref{sec: ermc} to improve its efficiency. Section~\ref{sec: exp} shows the experimental results. Section~\ref{sec: conclusion} concludes the article.

\section{Related Work}\label{sec: related_work}

\subsection{Adversarial Attacks}
Techniques such as the Fast Gradient Sign Method \cite{goodfellow2014explaining} and Projected Gradient Descent (PGD) \cite{madry2018towards} exploit the local gradient details of the target model to craft attacks. Building on PGD, output diversified sampling \cite{tashiro2020diversity} utilizes an enhanced initialization approach to create varied initial positions. However, these methods often provide inaccurate robustness measurements due to incorrect hyper-parameter tuning and gradient masking. To address this, Auto Attack (AA) \cite{croce2020reliable} combines four attack techniques with adjusted step sizes. To evaluate robustness under diversified $\ell_p$ norm perturbations simultaneously, Multi Steepest Descent (MSD) \cite{maini2020adversarial} incorporates various perturbation models within each step of the projected steepest descent, producing an adversary with a comprehensive understanding of the perturbation region. In this work, we consider PGD, AA, and MSD attacks to generate diversified $\ell_p$ norm perturbations.

\subsection{Adversarial Training-Based Defense}
The defense approach known as Adversarial Training (AT) \cite{madry2018towards} pioneered the use of min-max optimization for adversarial defense and has subsequently given rise to a plethora of other effective defense strategies. This includes the TRADES which delves into the trade-off between robustness and accuracy \cite{zhang2019theoretically}, dynamic adversarial training \cite{wang2019convergence}, and semi-supervised robust training approaches \cite{stanforth2019labels}. Furthermore, recent works, such as those by \cite{shafahi2019adversarial,wang2020fast,Wong2020Fast,zhang2019you}, have sought to develop faster, albeit approximate, AT algorithms. However, a common challenge across many of these methods is their concentration on a singular type of $\ell_p$ norm perturbation during AT. This specificity often culminates in a substantial decline in robustness when models are exposed to inputs with perturbations differing from the training set \cite{tramer2019adversarial}.

\subsection{Defenses on Diversified $\ell_p$ Norm Perturbations} 

Among all the works, \cite{croce2019provable} is the only one that provides a provable defense, and \cite{stutz2020confidence} considers withholding specific inputs to improve model resistance to stronger attacks. \cite{tramer2019adversarial} designs the inner loss by either selecting the type of perturbation that provides the maximum loss or averaging the loss across all types of perturbations. Extreme Norm Adversarial Training (E-AT) \cite{croce2022adversarial} leverages a fine-tuning strategy to improve robustness, while Multi Steepest Descent (MSD) Defense \cite{maini2020adversarial} incorporates various perturbation models within each step of the projected steepest descent to achieve diversified $\ell_p$ robustness. Nevertheless, despite their efforts, all the aforementioned works still depend on optimizing a single set of parameters, and the challenge of addressing the deficiency in diversified $\ell_p$ robustness remains unresolved. This work solves the challenge from a population-based optimization perspective.

\subsection{Population-Based Neural Network Learning} 
Optimizing a population of neural networks instead of a single network can prevent getting stuck at local minimums and lead to improved results. In one approach, \cite{jaderberg2017population} trained multiple instances of a model in parallel and selected the best performing instances to breed new ones. \cite{cui2018evolutionary} proposed an evolutionary stochastic gradient descent method that improved upon existing population-based methods. However, such methods typically have low learning speed and neglect adversarial robustness. Inspired by the human immune system, researchers have mimicked the key principles of the immune system in the inference phase to increase the robustness and not affect the learning speed in the training phase \cite{wang2022rails}. Mode connectivity can be treated as a faster population-based learning with two ancestor models that enhances the learning efficiency in the training phase \cite{garipov2018loss}. Researchers also analyze the mode connectivity when
networks are tampered with backdoor or error-injection attacks or under the attack of a single type of perturbation \cite{zhaobridging}. Our research extends beyond the scope of \cite{zhaobridging} by developing a novel robust, population-based optimization method for identifying candidate models with diversified $\ell_p$ robustness, and by exploring the phenomena of robust mode connectivity among different types of $\ell_p$ perturbations. Our unique approach not only facilitates the discovery of robust paths between two adversarially trained models but also generates candidates with enhanced robustness, thereby achieving state-of-the-art results in Diversified $\ell_p$ robustness. \textit{Our prior works laid the foundation for this study}: The workshop paper \cite{wang2023exploring} introduced Phase I robust path discovery but offered only limited experiments and lacked theoretical support, while the subsequent Deep Adversarial Defense \cite{wang2024deep} presented ERMC with preliminary validation. In this paper, we unify and extend these ideas into a full two-phase RMC framework, supported by SRMC, RMC-Based Optimization, refined ERMC, theoretical guarantees, and extensive experiments across datasets and architectures.

\section{Preliminaries}\label{sec: pre}

\subsection{Adversarial Attack with Different Input Perturbation Generators}
 
Recent studies indicate that conventional learning methods struggle with perturbed datasets ($\mathcal D_1,\mathcal D_2,\cdots, \mathcal D_S$) generated by
\begin{align}\label{eq: adv_atk}
    \displaystyle \arg\max {\mathcal L(\boldsymbol \theta; {\bf x}^\prime, y)}, ~~ s.t.~~ \text{Dist}_i({\bf x}^\prime, {\bf x}) \leq \delta_i, i \in [S]
\end{align}
for $\forall {\bf x} \in \mathcal D_0$, where $\mathcal D_0$ denotes the benign dataset, and $\delta_i$s are predefined bounds of perturbations corresponding to $\mathcal D_i$s with $i \in [S]$ (where $[S]=\{1, 2, \cdots, S\}$). In this paper, we restrict distance measures $\text{Dist}_i$s to be $\ell_p, p=1,2,\infty$ norms in our experiments. \eqref{eq: adv_atk} is typically termed an adversarial attack \cite{madry2018towards}. A practical approach to solving \eqref{eq: adv_atk} involves applying gradient descent and projection $P_{\delta_i}$ that maps the perturbation $\boldsymbol \epsilon_i=\bf x^\prime - \bf x$ to a feasible set, commonly referred to as a PGD attack. We use $\ell_p$-PGD to denote the PGD attack using the $\ell_p$ norm.

\subsection{Adversarial Training (AT)} 
The min-max optimization-based adversarial training (AT) is known as one of the most powerful defense methods to obtain a robust model against adversarial attacks \cite{madry2018towards}. We summarize AT below: 
\begin{align}
\begin{array}{ll}
    \displaystyle \min_{\boldsymbol \theta} \mathbb E_{(\mathbf x, y) \in \mathcal D_0} [ \displaystyle \max_{\text{Dist}_i({\bf x}^\prime, {\bf x}) \leq  \delta_i} 
 \mathcal L( \boldsymbol \theta; \mathbf x^\prime,  y ) ],
 \end{array}
 \label{eq: at}
\end{align}
Although AT can achieve relatively high robustness on $\mathcal D_i$, it does not generalize to other $\mathcal D_j, j\not=i$. Moreover, training on all $\mathcal D_i, i \in [S]$ is not scalable and will not provide robustness to all types of perturbations \cite{tramer2019adversarial}. We will use $\ell_p$-AT to denote the AT with the $\ell_p$ norm.

\subsection{Metric Definition: Diversified $\ell_p$ Robustness}

We hope that models can be robust to every $\ell_p$ adversarial type in the adversarial set of concerns, and we need to give a metric to measure such robustness. Diversified $\ell_p$ Robustness (DLR) is defined as its capacity to sustain the worst type of perturbation confined by a specific level of attack power:
\begin{mydef}\label{def_gr}

For a loss function $\mathcal L$, an input-output mapping function $f(\cdot)$, and a benign dataset $\hat{\mathcal D}_0$, the Diversified $\ell_p$ Robustness of a set of neural network parameters $\boldsymbol \theta$ is 
\begin{equation}\label{general_robustness_def}
\min_{i \in [S]} \frac{\sum_{({\bf x}^\prime, y)\in \hat{\mathcal D}_i} \boldsymbol{1}_{f({\bf x}^\prime, \boldsymbol \theta)= y}}{|\hat{\mathcal D}_i|},
\end{equation}
\end{mydef}
\noindent where $\hat{\mathcal D}_i$ represents the data generated by \eqref{eq: adv_atk} from $\hat{\mathcal D}_0$ with $\textup{Dist}_i$ as one of $\ell_p, p=1,2,\infty$ norms. We remark that there are other ways to define DLR. For example, the definition can be based on the worst-case sample-wise instead of worst-case data set-wise. Despite the difference, they essentially measure the same quantity, i.e., how well the model performs under various types of $\ell_p$ perturbations.

\subsection{Nonlinear Mode Connectivity}

\begin{figure}[h]
  \centering
  \includegraphics[trim=0 0 0 0,clip,width=.3\textwidth]{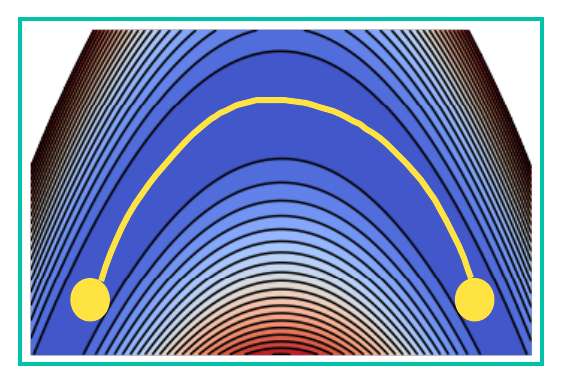}
  \caption{{The path with near-constant loss found by mode connectivity in the parameter space}. The endpoints are two pre-trained models, and any point on the path represents a model.} 
  \label{fig: mc_path}
\end{figure}

Mode connectivity is a neural network's property that local minimums found by gradient descent methods are connected by simple paths belonging to the parameter space \cite{freeman2017topology,garipov2018loss}. Everywhere on the paths achieves a similar cost as the endpoints. The endpoints are two sets of neural network parameters $\boldsymbol \theta_1, \boldsymbol \theta_2 \in \mathbb{R}^{d}$ with the same structure and trained by minimizing the given loss $\mathcal L$. The smooth parameter curve is represented using $\phi(t; \boldsymbol \theta) \in \mathbb{R}^{d}, t \in [0,1]$, such that $\phi(0; \boldsymbol \theta)=\boldsymbol \theta_1, \phi(1; \boldsymbol \theta)=\boldsymbol \theta_2$. To find a desired low-loss path between $\boldsymbol \theta_1$ and $\boldsymbol \theta_2$, it is proposed to find parameters that minimize the following expectation over a uniform distribution on the curve:
\begin{equation}
\min_{\boldsymbol \theta} \mathbb E_{t \sim q(t; \boldsymbol \theta)}  \displaystyle \mathbb E_{({\bf x},y) \sim \mathcal D_0}  \mathcal L( \phi(t; \boldsymbol \theta); ({\bf x},y)),
\label{eq: mc_original}
\end{equation}
where $q(t; \boldsymbol \theta)$ represents the distribution for sampling the parameters on the path. Note that \eqref{eq: mc_original} is generally intractable. A computationally tractable surrogate is proposed as follows 
\begin{equation}
\min_{\boldsymbol \theta} \mathbb E_{t \sim U(0,1)}  \displaystyle \mathbb E_{({\bf x},y) \sim \mathcal D_0}  \mathcal L( \phi(t; \boldsymbol \theta); ({\bf x},y)),
\label{eq: mc}
\end{equation}
where $U(0,1)$ denotes the uniform distribution on $[0,1]$. Two common choices of $\phi(t; \boldsymbol \theta)$ in nonlinear mode connectivity are the Bezier curve \cite{farouki2012bernstein} and Polygonal chain \cite{gomes2012computer}. As an example, a quadratic Bezier curve is defined as $\phi(t; \boldsymbol \theta) = (1-t)^2 \boldsymbol \theta_1 + 2t(1-t)\boldsymbol \theta + t^2 \boldsymbol \theta_2$. Training neural networks on these curves provides many similar-performing models on low-loss paths. As shown in Fig.~\ref{fig: mc_path}, a quadratic Bezier curve obtained from \eqref{eq: mc} connects the upper two models along a path of near-constant loss.

\section{Two-Phase Robust Mode Connectivity}

\subsection{Phase I: Robust Path Search Via Robust Mode Connectivity}\label{sec: rmc}

\subsubsection{A Pilot Exploration}
Mode connectivity and adversarial training seem to be two excellent ideas for achieving high DLR that has been defined in Definition~\ref{def_gr}. If we set $\phi(0; \boldsymbol \theta)$ and $\phi(1; \boldsymbol \theta)$ to be two adversarially-trained neural networks under different types of perturbations, applying \eqref{eq: mc} may result in a path with points having high robustness for all these perturbations. Thus we ask:
\begin{tcolorbox}[before skip=2.0mm, after skip=2.0mm, boxsep=0.0cm, middle=0.1cm, top=0.1cm, bottom=0.1cm]
\noindent \textit{\textbf{(Q1)} Can simply combining adversarial training with mode connectivity provide high DLR?}
\end{tcolorbox}

\begin{figure}[h]
  \centering
  \includegraphics[trim=0 0 0 0,clip,width=.46\textwidth]{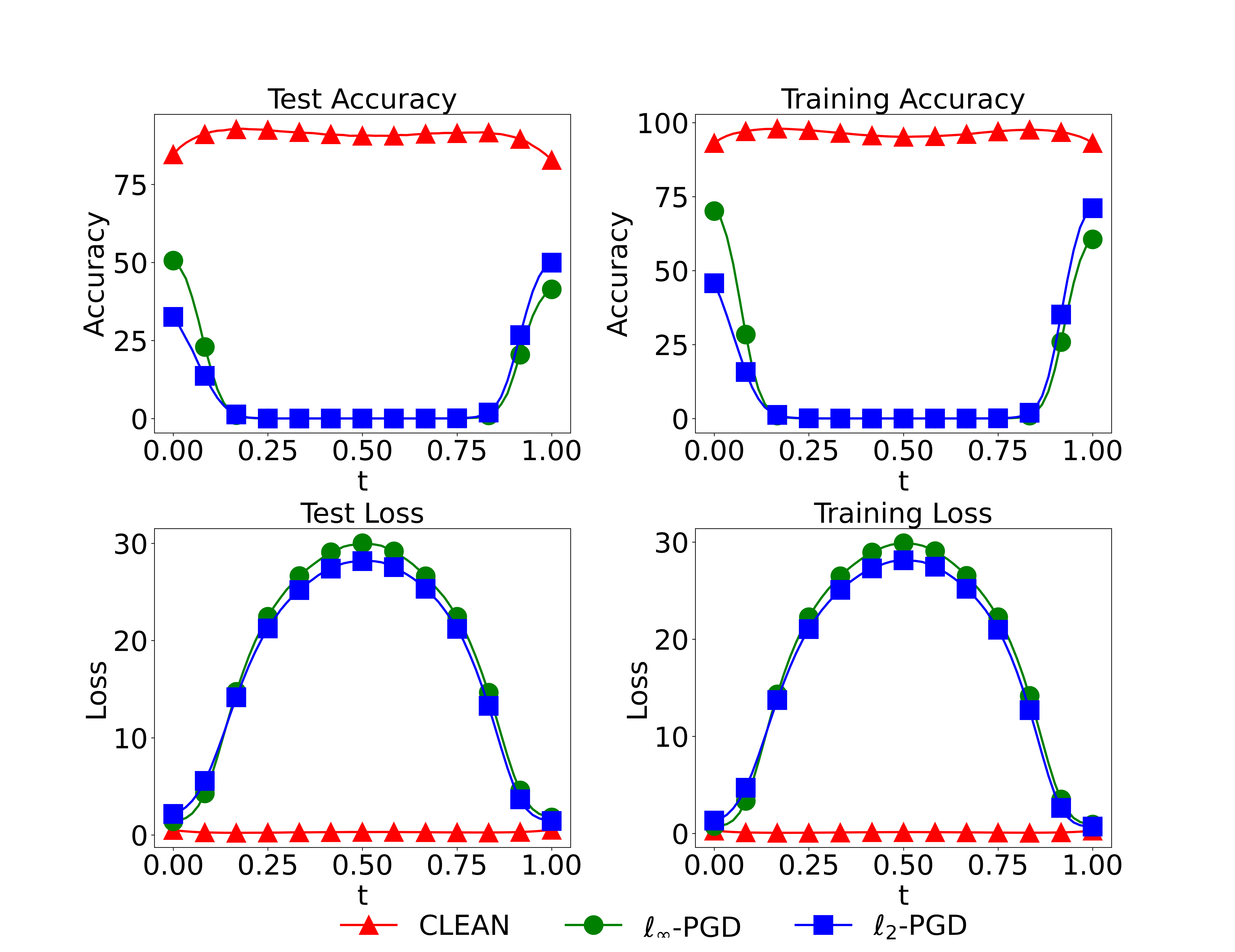}
  \caption{{The vanilla mode connectivity \eqref{eq: mc} with models trained by $\ell_\infty$-AT and $\ell_2$-AT as two endpoints fails to find the path with high DLR}. $\phi(0; \boldsymbol \theta)$ and $\phi(1; \boldsymbol \theta)$ are $\ell_\infty$-AT ($\delta=8/255$, 150 epochs) and $\ell_2$-AT ($\delta=1$, 150 epochs). } 
  \label{fig: std_mc}
\end{figure}

Here we aim to see if implementing vanilla mode connectivity can bring us high DLR. We combine two PreResNet110 models \cite{he2016identity}, one trained with $\ell_\infty$-AT ($\delta=8/255$, 150 epochs) and the other trained with $\ell_2$-AT ($\delta=1$, 150 epochs), to find the desired path using the vanilla mode connectivity \eqref{eq: mc}. $\phi(0; \boldsymbol \theta)$ and $\phi(1; \boldsymbol \theta)$ are models trained by $\ell_\infty$-AT and $\ell_2$-AT, respectively. The mode connectivity curve is obtained with additional 50 epochs of training. The results are shown in Fig.~\ref{fig: std_mc}. The left (right) endpoint represents the model trained with $\ell_\infty$-AT ($\ell_2$-AT). One can see that the path has high loss and low robust accuracies on both types of attacks, indicating that vanilla mode connectivity fails to find the path that enjoys high DLR.

\subsubsection{Embedding Adversarial Robustness to Mode Connectivity}

Although the vanilla mode connectivity aims to provide insight into the loss landscape geometry, it searches space following the original data distribution. Therefore it cannot provide high DLR by simply using two adversarially-trained models as two endpoints. Instead, we ask:
\begin{tcolorbox}[before skip=2.0mm, after skip=2.0mm, boxsep=0.0cm, middle=0.1cm, top=0.1cm, bottom=0.1cm]
\noindent \textit{\textbf{(Q2)} Can we develop a new method to embed adversarial robustness to mode connectivity by searching the adversarial input space?}
\end{tcolorbox}

To answer (Q2), we connect mode connectivity \eqref{eq: mc} with adversarial training under diversified $\ell_p$ adversarial perturbations. In other words, we modify the objective \eqref{eq: mc} to adjust to our high DLR purpose. An adversarial generator is added as an inner maximization loop. We adopt different types of perturbations in the generator. This is because a single type of perturbation may result in robustness bias. Formally, we obtain a model path $\phi(t; \boldsymbol \theta), t \in [0,1]$ parameterized by $\boldsymbol \theta$.  
\begin{equation}
\min_{\boldsymbol \theta} \mathbb E_{t \sim U(0,1)}  \displaystyle \mathbb E_{({\bf x},y) \sim \mathcal D_0 } \sum_{i \in I} \max_{\text{Dist}_i({\bf x}^\prime, {\bf x}) \leq  \delta_i} \mathcal L( \phi(t; \boldsymbol \theta); ({\bf x}^\prime,y)),
\label{eq: mc_adv}
\end{equation}
where $\phi(0; \boldsymbol \theta)$ and $\phi(1; \boldsymbol \theta)$ are two models trained by \eqref{eq: at}, probably under different types of perturbations. Throughout this paper, we fix the curve as a quadratic Bezier curve. Thus a model at the point $t$ can be represented by $\phi(t; \boldsymbol \theta) = (1-t)^2 \boldsymbol \theta_1 + 2t(1-t)\boldsymbol \theta + t^2 \boldsymbol \theta_2$. Similar to the nonlinear mode connectivity, \eqref{eq: mc_adv} is a computationally tractable relaxation by directly sampling $t$ from the uniform distribution $U(0,1)$ during the optimization. Data points $({\bf x}^\prime,y)$ are generated from a union of adversarial strategies $i \in I$, where $I$ is a subset of $\{1,2, 3, \cdots, S\}$. For example, data can be generated by using $\ell_2$ or $\ell_\infty$ norm distance measure, which is commonly used in adversarial attacks and adversarial training. Formulation in \eqref{eq: mc_adv} ensures that the identified path adapts to the targeted adversarial perturbations. 

We call \eqref{eq: mc_adv} the Robust Mode Connectivity (RMC), which serves as the first defense phase for robust path search. We remark that RMC itself is a defense method as we can select the model with the highest DLR in the path. One can see that a group of models (all points in the path) are generated from two initial models. Therefore RMC is a population-based optimization. 

\subsubsection{On the Benefits of Population-Based Framework}
Training a single model presents a fundamental challenge: different data points, and even the same point under varying adversarial norms, do not achieve peak robustness simultaneously. We therefore ask: 
\begin{tcolorbox}[before skip=2.0mm, after skip=2.0mm, boxsep=0.0cm, middle=0.1cm, top=0.1cm, bottom=0.1cm]
\noindent \textit{\textbf{(Q3)} Can a population-based framework address this issue by leveraging a large number of models?}
\end{tcolorbox}
Here, we define ``simultaneous'' robustness at the epoch-level, i.e., data points are considered to peak concurrently if they do so within the same training epoch. We posit that with a sufficiently large population of models, it is possible for $N$ data points to achieve peak robustness simultaneously under $S$ distinct adversarial norms.

\begin{theorem}
Let $T\geq 2$ be the number of training epochs. For each model $k$, each data-point/robustness-type pair $(i,s) \in \{1,\cdots, N\}\times \{1,\cdots, S\}$ has a random variable $X_{i,s}^{(k)} \in \{1,\cdots, T\}$ equal to the epoch at which that pair attains its highest $\ell_p$ robustness in model $k$. Assume the arrays $\{X_{i,s}^{(k)}\}_{i,s}$ are i.i.d. over $k$, the $X_{i,s}^{(k)}$ are mutuallly independent across $(i,s)$, and $\text{Pr}[X_{i,s}^{(k)}=t]=\frac{1}{T}$ for every $t \in \{1,\cdots, T\}$ for all $(i,s)$, then:

\noindent For any target confidence $1-\gamma \in (0,1)$, the minimum number of models that achieves $Pr$(at least one alignment among the $K$ models)$\geq 1-\gamma$ is $K=\lceil{\frac{\ln{\gamma}}{\ln{1-T^{-(NS-1)}}}\rceil}$.
\end{theorem}
The alignment event in each model has the probability $T^{-(NS-1)}$, so the probability of no alignment in the $K$ models is $1-T^{-(NS-1)}$. The proof is done by requiring $(1-T^{-(NS-1)})^K\le \gamma$. Note that the independence assumption may not hold in practice. Nevertheless, our objective here is solely to demonstrate the inherent advantages of the population-based method over a single model solution.

With the problem formulation and the theoretical support, the next step is to find out how to solve the RMC \eqref{eq: mc_adv}.

\begin{algorithm}[h]
\caption{Robust Mode Connectivity}
\label{alg: RMC}
\begin{algorithmic}[1]
\REQUIRE $\phi(0; \boldsymbol \theta)$, $\phi(1; \boldsymbol \theta)$ - two selected models with the same structure (potentially trained with different strategies, e.g., AT under different perturbation types); initial model $\boldsymbol \theta^0$; the perturbation types $i \in I$ and the corresponding projections $P_{\delta_i}$; training set $\mathcal D_0$; inner loop iteration number $J$; batch size $B$; initial perturbation $\epsilon^{(0)}=\mathbf 0$.
\STATE{$\boldsymbol \theta = \boldsymbol \theta^0$}
\FOR{each data batch $\mathcal D_b \in \mathcal D_0$ in each epoch $e \in E$}
\STATE{Uniformly select $t \sim U(0,1)$}
\FOR{$\forall \bf x \in \mathcal D_b$}
\FOR{$j = 1, \cdots, J$}
\FOR{$i \in I$}
\STATE{\hspace{-1mm}$\boldsymbol{\epsilon}_i^{(j)} \leftarrow P_{\delta_i}\big(\boldsymbol{\epsilon}^{(j-1)}-{\nabla_{\boldsymbol{\epsilon}}\mathcal L(\phi(t; \boldsymbol{\theta}); {\bf{x}}+{\boldsymbol{\epsilon}}^{(j-1)},y)}\big)$}
\ENDFOR
\STATE{$\boldsymbol \epsilon^{(j)} \leftarrow \arg\max_{\boldsymbol \epsilon_i^{(j)}, i \in I} {\mathcal L(\phi(t; \boldsymbol \theta); {\bf{x}} + \boldsymbol \epsilon_i^{(j)}, y)}$}
\ENDFOR
\ENDFOR
\STATE{$\boldsymbol \theta \leftarrow \boldsymbol \theta - \nabla_{\boldsymbol \theta} \sum_{\bf x \in \mathcal D_b} \mathcal L(\phi(t; \boldsymbol \theta); {\bf x} + \boldsymbol \epsilon^{(j-1)}, y)$}
\ENDFOR
\RETURN $\boldsymbol \theta$, $\phi(t; \boldsymbol \theta), \forall t \in [0,1]$
\end{algorithmic}
\end{algorithm}

\begin{figure*}[ht]
  \centering
             \subfloat[Epoch=50]{\includegraphics[trim=90 5 90 100,clip,width=0.32\textwidth]{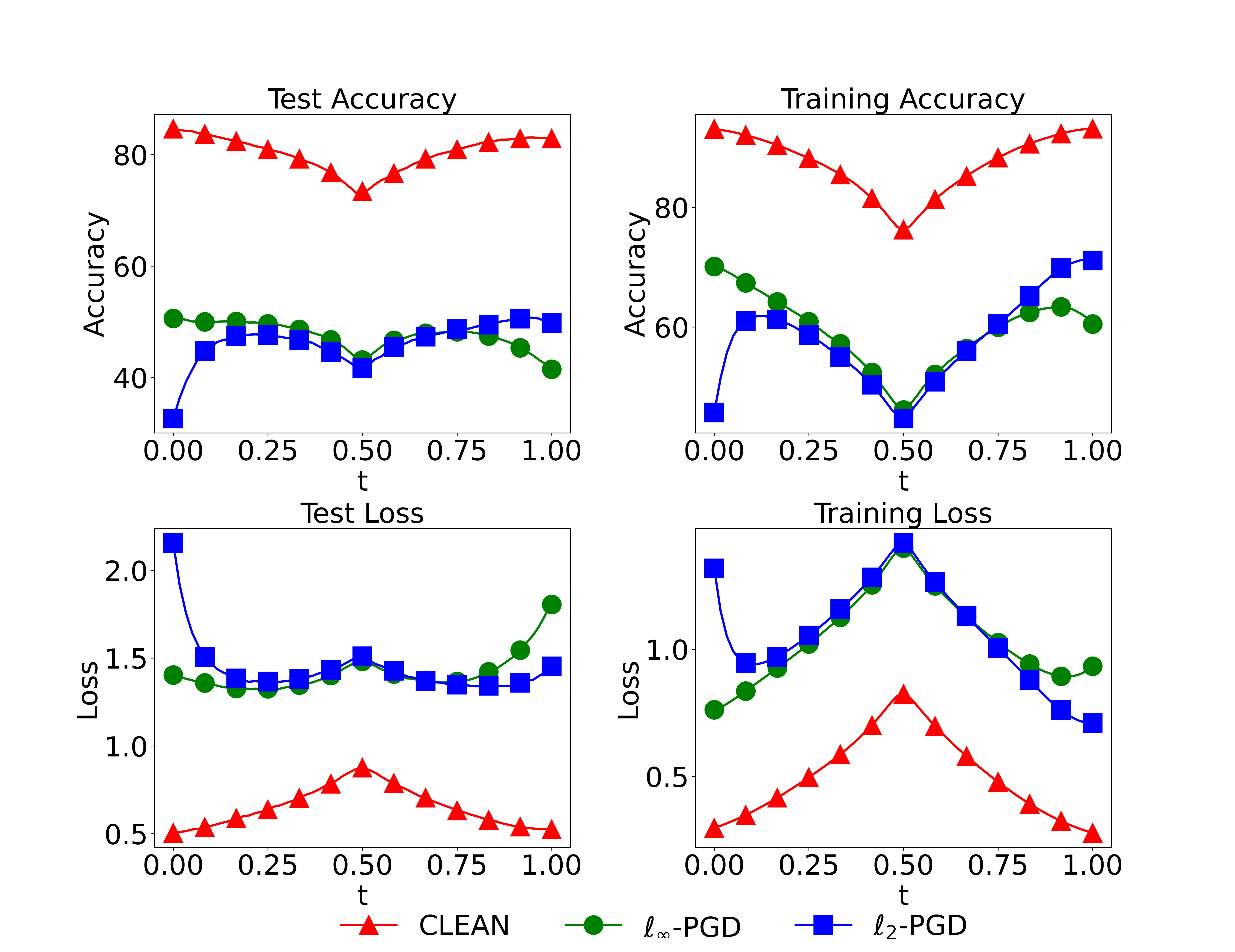}}
             \subfloat[Epoch=100]{\includegraphics[trim=90 5 90 100,clip,width=0.32\textwidth]{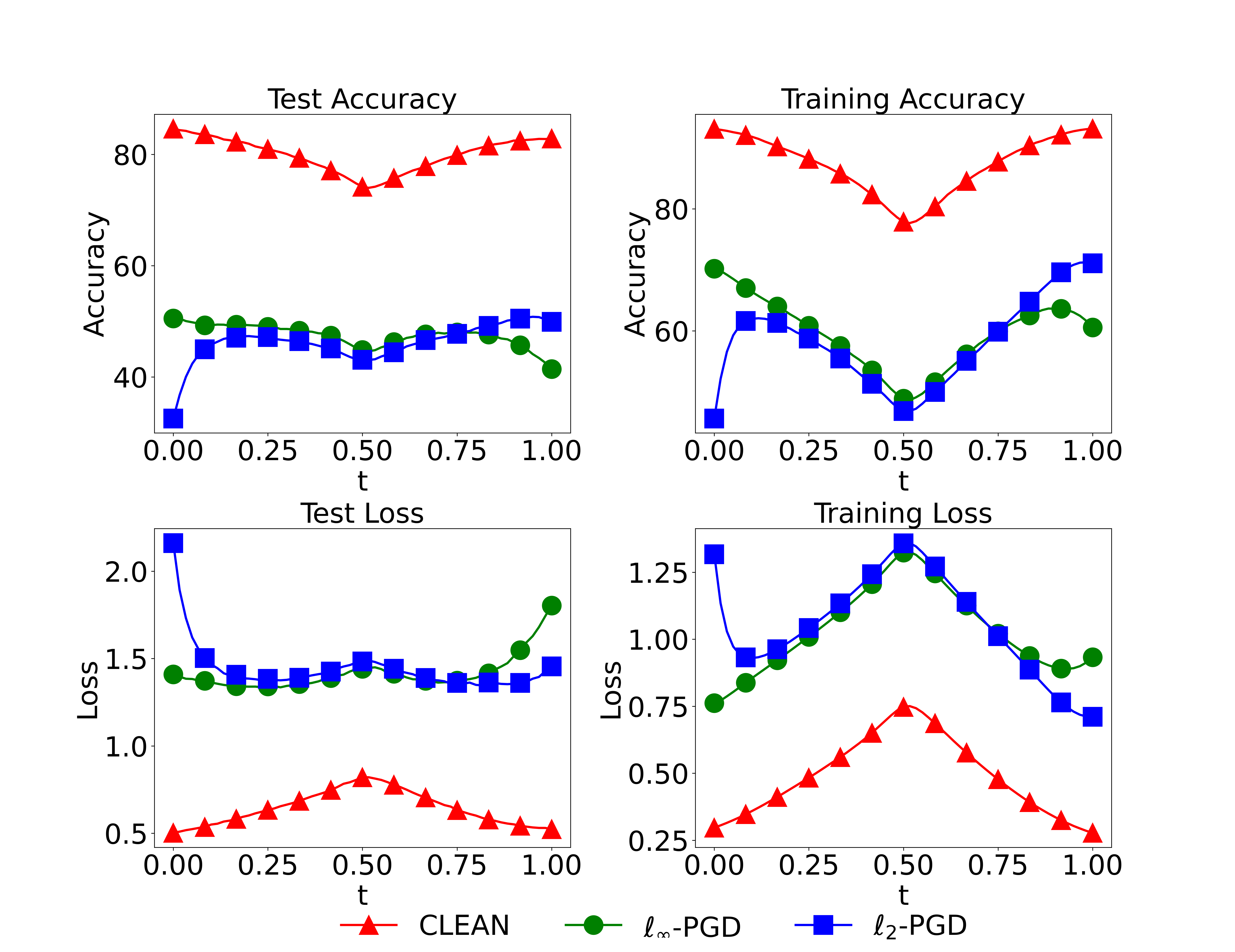}}
             \subfloat[Epoch=150]{\includegraphics[trim=90 5 90 100,clip,width=0.32\textwidth]{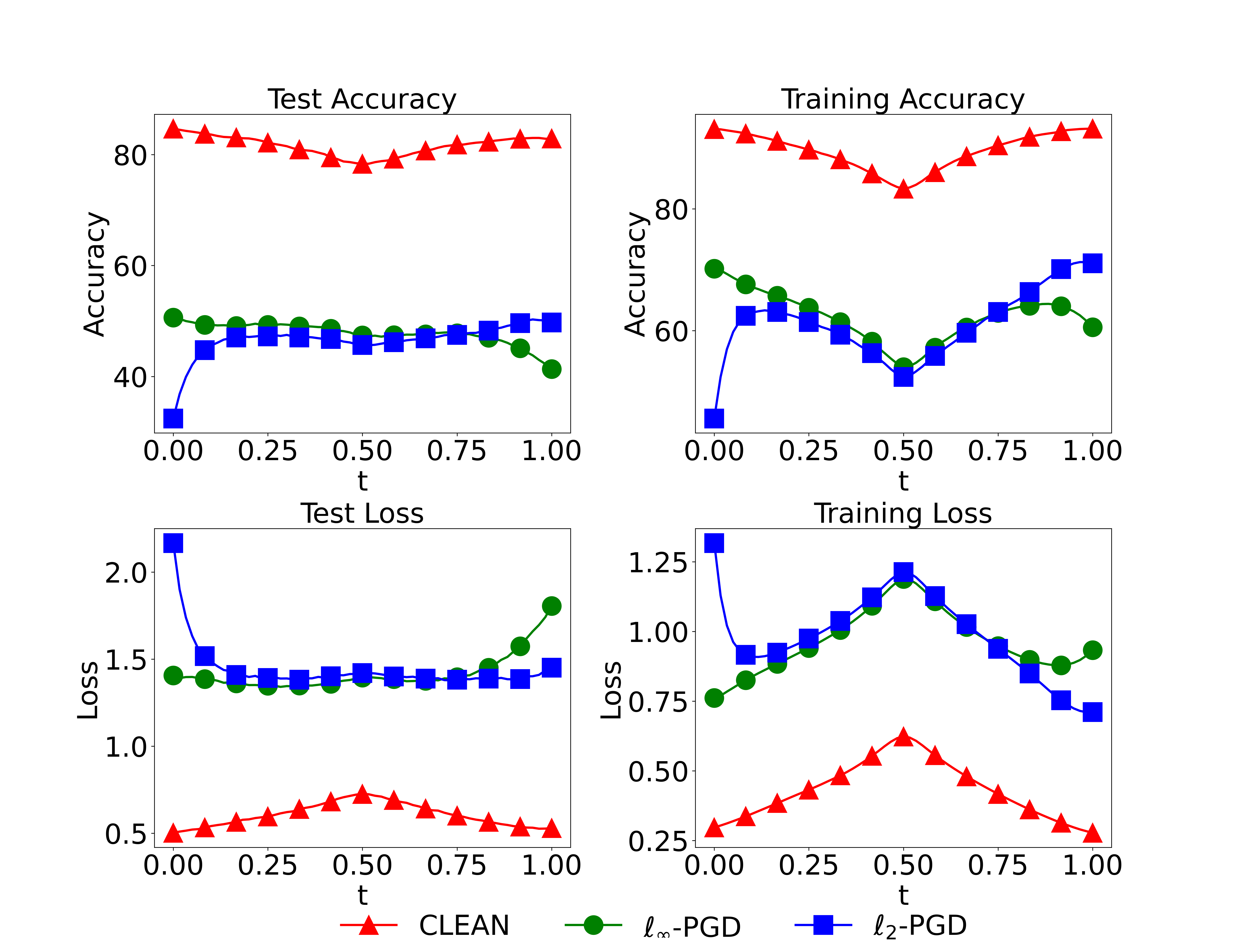}}
  \caption{{The RMC \eqref{eq: mc_adv} with models trained by $\ell_\infty$-AT and $\ell_2$-AT as two endpoints can find the path with high DLR}. MSD \cite{maini2020adversarial} with perturbations generated by $\ell_2$ and $\ell_\infty$ norm distance measures is leveraged as the inner solver. Solving \eqref{eq: mc_adv} uses 50/100/150 epochs in panel (a)/(b)/(c).} 
  \label{fig: adv_mc}
\end{figure*}

\subsubsection{Solving Robust Mode connectivity Via Multi Steepest Descent}

Solving \eqref{eq: mc_adv} is difficult as it contains multi-type perturbations. The simplest ways are using `MAX' or `AVG' strategy proposed in \cite{tramer2019adversarial}, where the inner loss is obtained by selecting the type of perturbation that provides the maximum loss or averaging the loss on all types of perturbations. However, both strategies consider perturbations independently. We leverage a Multi Steepest Descent (MSD) approach that includes the various perturbation models within each step of the projected steepest descent in order to produce a PGD adversary with complete knowledge of the perturbation region \cite{maini2020adversarial}. The key idea is to simultaneously maximize the worst-case loss overall perturbation models at each step. Algorithm~\ref{alg: RMC} shows the details, where all the perturbation types are considered in each iteration. The complexity order remains consistent when we juxtapose the Algorithm with the conventional AT. This is primarily because the number of perturbations $I$ is essentially constant in our scenarios (specifically, $I=2$ or $3$). Next we test the effectiveness of the proposed RMC algorithm.

We again use models trained with $\ell_\infty$-AT ($\delta=8/255$, 150 epochs) and $\ell_2$-AT ($\delta=1$, 150 epochs) as two endpoints $\phi(0; \boldsymbol \theta)$ and $\phi(1; \boldsymbol \theta)$. The RMC \eqref{eq: mc_adv} with MSD as the inner solver is applied to obtain the path. Fig.~\ref{fig: adv_mc} shows the results of training an additional 50/100/150 epochs with $D_{i}$s generated by $\ell_2$ and $\ell_\infty$ norm distance measures. One can find that unlike Fig.~\ref{fig: std_mc}, the paths contain points with high accuracy and high robustness against both $\ell_\infty$-PGD and $\ell_2$-PGD attacks. Although the left endpoint has low $\ell_2$ robustness and the right endpoint has relatively low $\ell_\infty$ robustness, they can achieve high DLR in the connection, where the highest DLR is $48.19\%$ in panel (a). One can also notice that when the epoch number for solving \eqref{eq: mc_adv} increases, the path becomes smoother. One can see that the robust paths also function as effective mode connectivity paths, where both the clean accuracy and loss (indicated by red lines) maintain consistent levels between the two endpoints $t=0$ and $t=1$ in panels (c). We also observe that the optimal points in panels (a), (b), and (c) yield similar DLR. The experiments indicate that RMC can find a path with high DLR. If the goal is to select an optimal model from the path, it is enough to only conduct the training with a small number of epochs.

\subsubsection{Improving Learning Efficiency of the RMC With a Self-Robust Mode Connectivity Module}\label{sec: srmc}

One drawback of the previous scheme is that it needs to initially pre-train two neural networks, which could be slow when the computational resources are limited. We ask:
\begin{tcolorbox}[before skip=2.0mm, after skip=2.0mm, boxsep=0.0cm, middle=0.1cm, top=0.1cm, bottom=0.1cm]
\noindent \textit{\textbf{(Q4)} How can we accelerate the learning process of the RMC?}
\end{tcolorbox}

Here we propose to replace RMC with a self-robust mode connectivity (SRMC) module in the learning process. SRMC can accelerate the endpoints training in the path search process and thus speed up RMC. We start with one set of neural network parameters $\phi(0; \boldsymbol \theta)=\boldsymbol \theta_1$ that is trained by \eqref{eq: at} with a fixed $\text{Dist}_i$. After the model achieves high robustness on $\mathcal D_i$, we retrain the model for a few epochs using \eqref{eq: at} with $\text{Dist}_j$. The new model we obtained will be placed at the endpoint $\phi(1; \boldsymbol \theta)=\boldsymbol \theta_2$. Now the low-loss high-robustness path can be found by optimizing \eqref{eq: mc_adv}. By leveraging the SRMC module, our proposed framework yields both high DLR and learning efficiency.

\subsection{Phase II: Robust Model Selection Via Robust Mode Connectivity-Based Optimization}\label{sec: opt}

\begin{algorithm}[h]
\caption{Robust Mode Connectivity-Based Optimization ($\ell_1, \ell_2, \ell_\infty$ perturbations)}
\label{alg: RMC_opt}
\begin{algorithmic}[1]
\STATE{Train three models for $T$ epochs using $\ell_1, \ell_2, \ell_\infty$ perturbations, respectively. (Training can be accelerated using the SRMC module proposed in Section~\ref{sec: srmc})}
\STATE{Apply Algorithm~\ref{alg: RMC} with $\ell_2, \ell_1$-AT trained models ($I$ includes $\ell_2, \ell_1$) and $\ell_\infty, \ell_1$-AT trained models ($I$ includes $\ell_\infty, \ell_1$) as $\phi(0; \boldsymbol \theta)$, $\phi(1; \boldsymbol \theta)$, and return model trajectories $\phi_{\boldsymbol \theta-\ell_\infty}(t), \phi_{\boldsymbol \theta-\ell_2}(t), \forall t \in [0,1]$. (pairs of perturbations can be selected in different ways)}
\STATE{Randomly pick points $t_{-\ell_\infty}$, $t_{-\ell_2}$ from  optimal regions for each model trajectory.}
\STATE{Train models for $T$ epochs using $\ell_\infty, \ell_2$ perturbations starting from $\phi_{\boldsymbol \theta-\ell_\infty}(t_{-\ell_\infty}), \phi_{\boldsymbol \theta-\ell_2}(t_{-\ell_2})$, respectively.}
\STATE{Apply Algorithm~\ref{alg: RMC} with the two models as $\phi(0; \boldsymbol \theta)$, $\phi(1; \boldsymbol \theta)$ with $I$ including $\ell_1, \ell_2, \ell_\infty$ perturbations.}
\STATE{Find the optimal point $t_{\text{opt}}$ from the model trajectory.}
\RETURN $\phi_{\boldsymbol \theta}(t_{\text{opt}})$
\end{algorithmic}
\end{algorithm}

\subsubsection{From RMC to RMC-Based Optimization}

Suppose we have neural networks that share the same structure but are trained with different settings, e.g., different types of perturbations, perturbation magnitudes, learning rates, batch size, etc. In that case, we can use RMC to search for candidates potentially leading us to better solutions or even global optimums. The intuition behind the claim is that low-loss \& high-DLR paths connect all the minimums, and thus it becomes easier for search algorithms to jump out of the sub-local minimums. We have seen the exciting property of the proposed RMC, which indicates that a larger population of NNs can result in higher DLR. Notice that RMC can serve as a component in a larger population-based optimization to select robust models with higher DLR. A natural question to ask is:
\begin{tcolorbox}[before skip=2.0mm, after skip=2.0mm, boxsep=0.0cm, middle=0.1cm, top=0.1cm, bottom=0.1cm]
\noindent \textit{\textbf{(Q5)} Can we develop a general population-based optimization method built on RMC modules to further improve the DLR of a single RMC?}
\end{tcolorbox}

The RMC-based optimization we developed below includes an evolving process of RMC units for multiple generations. As a starting point, we generate an initial population by training neural networks with data points augmented using different $\text{Dist}_i$ in \eqref{eq: at}. 
We use gradient descent to train these networks but pause the training when specific stop criteria have been met, e.g., the number of epochs or accuracy achieving the preset threshold. The initial population then varies, and the system selects candidates with the best performances. The two operations in our approach are unified through the RMC that connects two adversarially-trained neural network models on their loss landscape using a high-accuracy \& high-DLR path characterized by a simple curve. Candidates for the next generation are selected among the high DLR points on the curve. The process can be repeated and an optimal solution that enjoys the highest DLR is selected among the final candidates. 

Algorithm~\ref{alg: RMC_opt} shows the pipeline using an example of three types of perturbations. We first train three models with $\ell_\infty$-AT, $\ell_2$-AT, and $\ell_1$-AT for $T$ epochs. We then connect the $\ell_2$-AT model with the $\ell_1$-PGD model and connect the $\ell_\infty$-AT model with the $\ell_1$-AT model using the RMC for some additional epochs. The two model trajectories are denoted by $\phi_{\boldsymbol \theta-\ell_\infty}(t)$ and $\phi_{\boldsymbol \theta-\ell_2}(t), \forall t \in [0,1]$. Notice that the curves will not be perfectly flat. But there exist some regions containing points with high DLR. We will randomly pick a model from a small optimal region in each curve. In practice, we will find the point with the highest DLR and randomly pick a point around the optimal point. The rationale behind this is to increase diversity during the training. After obtaining the models $\phi_{\boldsymbol \theta-\ell_\infty}(t_{-\ell_\infty})$ and $ \phi_{\boldsymbol \theta-\ell_2}(t_{-\ell_2})$ from both trajectories, the two new endpoints are obtained by training each model $T$ epochs using the $\ell_p$-AT that is different from the types used in the previous RMC. In this specific case, we use $\ell_\infty$-AT and $\ell_2$-AT. Finally, we connect the two new endpoints with RMC for some additional epochs and find the new optimum $\phi_{\boldsymbol \theta}(t_{\text{opt}})$ at $t_{\text{opt}}$. In the case of two types of perturbations, one can start to train two models from a single optimal point or train one model from each of the two optimal points. We refer readers to Section~\ref{subsec: rmc_opt} for more details. It's pertinent to note that parameter curves derived from distinct models can be concurrently computed. This inherent parallelizability means that when we leverage parallel computing for generating independent parameter curves, the execution time is equivalent to the time of generating one parameter curve. In a more general scenario where there are $S$ types of perturbations, the process is the same, except that it contains more RMC units, as illustrated in Fig.~\ref{fig: framework}. We learn optimal points from pairs of models trained by AT under different perturbations and finally find an optimal point with the highest DLR.

\section{Enhancing Phase II efficiency: ERMC With Model Ensemble}\label{sec: ermc}

From the insights of the previous Phase II, it becomes apparent that to enhance robustness against diversified $\ell_p$ perturbations, multiple RMC procedures might be necessary. We pose the question:
\begin{tcolorbox}[before skip=2.0mm, after skip=2.0mm, boxsep=0.0cm, middle=0.1cm, top=0.1cm, bottom=0.1cm]
\noindent \textit{\textbf{(Q6)} Can enhanced robustness against diversified $\ell_p$ perturbations be attained within a single RMC process?}
\end{tcolorbox}

In the literature \cite{croce2019provable}, it is discussed that affine classifiers, including CNNs with ReLU activations, can withstand various $\ell_p$ norm attacks if they are already fortified against $\ell_1$ and $\ell_\infty$ perturbations. Theorem 3.1 in ~\cite{croce2019provable} posits that the convex hull of the union ball of the $\ell_1$ and $\ell_\infty$ provides satisfactory robustness to $\ell_p, 1 \le p\le \infty$ perturbationsTheorem 3.1 in ~\cite{croce2019provable} posits that the convex hull of the union ball of the $\ell_1$ and $\ell_\infty$ provides satisfactory robustness to $\ell_p$ perturbations, where $1 \le p\le \infty$. 

\begin{theorem}
    \cite{croce2019provable} Suppose that the classifier is affine. Let $C$ be the convex hull of the union ball of the $\ell_1$ and $\ell_\infty$. If the input dimension $d_{\mathbf x}$ is larger than or equal to two and $\delta_1 \in (\delta_\infty, d_{\mathbf x}\delta_\infty)$, then
    \begin{equation}
  \min_{\mathbb{R}^{d_{\mathbf x}}\backslash C} \|\mathbf x^\prime - \mathbf x\|_p = \frac{\boldsymbol \delta_1}{(\boldsymbol \delta_1/\boldsymbol \delta_\infty - \beta + \beta^q)^{1/q}}
    \end{equation}
where $\beta=\frac{\boldsymbol \delta_1}{\boldsymbol \delta_\infty} - \lfloor \frac{\boldsymbol \delta_1}{\boldsymbol \delta_\infty}\rfloor$ and $\frac{1}{p}+\frac{1}{q}=1$.
\end{theorem}

A recent approach, E-AT \cite{croce2022adversarial}, proposes using fine-tuning as an efficient transition from $\ell_\infty$-adversarial training (AT) to $\ell_1$-AT, aiming to improve robustness against a broader range of $\ell_p$ attacks. However, this method faces two key limitations: \ding{182} the fine-tuning process may compromise the model's original robustness to $\ell_\infty$ attacks; and \ding{183} a single model often lacks sufficient capacity to maintain strong robustness against both $\ell_\infty$ and $\ell_1$ perturbations simultaneously.

Our proposed RMC can naturally overcome these issues thanks to the power of population-based strategies. Here we propose an efficient robust mode connectivity (ERMC) strategy, leveraging SRMC to fine-tune endpoint $\phi(1; \boldsymbol \theta)$ with $\ell_1$-AT from $\phi(0; \boldsymbol \theta)$ obtained by $\ell_\infty$-AT. We then optimize the following objective to maintain robustness against both $\ell_1$ and $\ell_\infty$ attacks, effectively expanding the defense boundary and improving overall model resilience:  
\begin{equation}
\begin{aligned}
\min_{\boldsymbol \theta} &\mathbb E_{t \sim U(0,1)}   \displaystyle \mathbb E_{({\bf x},y) \sim \mathcal D_0 } \{\\&\sum_{\text{Dist}_i \in \{\|\cdot\|_1,\|\cdot\|_\infty\}} \max_{\text{Dist}_i({\bf x}^\prime, {\bf x}) \leq  \delta_i} \mathcal L( \phi(t; \boldsymbol \theta); ({\bf x}^\prime,y))\},
\label{eq: ermc_adv}
\end{aligned}
\end{equation} 
which results in a larger union ball, thereby enhancing the model's resilience against a broader range of perturbations. The detailed algorithm can be found in Algorithm~\ref{alg: ERMC}. ERMC is efficient as it only needs to conduct the connection once.

\begin{algorithm}[h]
\caption{Efficient Robust Mode Connectivity}
\label{alg: ERMC}
\begin{algorithmic}[1]
\REQUIRE A model $\phi(0; \boldsymbol \theta)$ trained with $\ell_\infty$-AT; initial model $\boldsymbol \theta^0$; the corresponding projections $P_{\boldsymbol{\delta}_1}$ and $P_{\boldsymbol{\delta}_\infty}$; training set $\mathcal D_0$; iteration number $J$; batch size $B$; initial perturbation $\boldsymbol{\delta}^{(0)}=\mathbf 0$.
\STATE{Create a copy of $\phi(0; \boldsymbol \theta)$ and retrain it with AT-$\ell_1$ for 10 epochs to obtain a model $\phi(1; \boldsymbol \theta)$.}
\STATE{$\boldsymbol \theta = \boldsymbol \theta^0$}
\FOR{each data batch $\mathcal D_b \in \mathcal D$ in each epoch $e \in E$}
\STATE{Uniformly select $t \sim U(0,1)$}
\FOR{$\forall \bf x \in \mathcal D_b$}
\FOR{$j = 1, \cdots, J$}
\STATE{\hspace{-1mm}$\boldsymbol{\delta}_1^{(j)} \leftarrow P_{\boldsymbol \epsilon_1}\big(\boldsymbol{\delta}^{(j-1)}-{\nabla_{\boldsymbol{\delta}}\mathcal L( \phi(t; \boldsymbol \theta); {\bf{x}}+{\boldsymbol{\delta}}^{(j-1)},y)}\big)$}
\STATE{\hspace{-1mm}$\boldsymbol{\delta}_\infty^{(j)} \leftarrow P_{\boldsymbol \epsilon_\infty}\big(\boldsymbol{\delta}^{(j-1)}-{\nabla_{\boldsymbol{\delta}}\mathcal L( \phi(t; \boldsymbol \theta); {\bf{x}}+{\boldsymbol{\delta}}^{(j-1)},y)}\big)$}
\ENDFOR
\STATE{$\boldsymbol \delta^{(j)} \leftarrow \arg\max_{\boldsymbol \delta_i^{(j)}, i \in \{1,\infty\}} {\mathcal L( \phi(t; \boldsymbol \theta); {\bf{x}} + \boldsymbol \delta_i^{(j)}, y)}$}
\ENDFOR
\STATE{$\boldsymbol \theta \leftarrow \boldsymbol \theta - \nabla_{\boldsymbol \theta} \sum_{\bf x \in \mathcal D_b} \mathcal L(\phi(t; \boldsymbol \theta); {\bf x} + \boldsymbol \delta^{(j)}, y)$}
\ENDFOR
\RETURN $\boldsymbol \theta$, $\phi(t; \boldsymbol \theta), \forall t \in [0,1]$
\end{algorithmic}
\end{algorithm}

Acknowledging the presence of numerous models along the trajectory that demonstrate significant resistance to $\ell_\infty$ and $\ell_1$ attacks, adopting a model ensemble technique seems a logical step to enhance robustness. By doing so, we create an aggregated model with a collective defense against both $\ell_\infty$ and $\ell_1$ disruptions. The process for selecting the ensemble involves identifying a segment $t\in [a,b]$ on the path $\phi(t; \boldsymbol \theta)$ where each point on the segment has robust accuracies surpassing two prefixed model selection thresholds $\alpha_\infty, \alpha_1$ under $\ell_\infty$ and $\ell_1$ attacks, respectively. From this segment, we select $n > 1$ models situated at intervals defined by $t = a + \frac{b - a}{n - 1}i$, with $i$ varying from $0$ to $n - 1$. In scenarios where there are several non-adjacent intervals that fulfill the selection criteria, the models are proportionally allocated based on the length of these intervals. This approach, with $n$ models chosen, is referred to as ERMC-$n$. We then calculate the class probability prediction by averaging the outputs from the final layers of these $n$ models.

\section{Experimental Results}\label{sec: exp}

Figures~\ref{fig: std_mc}, \ref{fig: adv_mc} show that using the proposed RMC can find a path with points in it enjoying high robustness on diversified $\ell_p$ perturbations. In this section, we conduct more comprehensive experiments on the Robust Mode Connectivity-Oriented Adversarial Defense.

\subsection{Settings}

We evaluate our proposed methods using the CIFAR-10, CIFAR-100 \cite{krizhevsky2009learning}, and ImageNet-100 \cite{russakovsky2015imagenet} datasets across the PreResNet110, WideResNet-28-10, and Vision Transformer-base architectures. By default, we conduct experiments on CIFAR-10 and PreResNet110. For our experiments, the considered perturbation types, denoted as $\textup{Dist}_i$s, are based on $\ell_\infty$, $\ell_2$, and $\ell_1$ norms, constrained by perturbations of $\delta = 8/255, 1$, and $12$, respectively. To obtain the endpoints' models, we employ AT. Our methods are benchmarked against the standard $\ell_\infty$-AT baseline, RMC on two randomly initialized models (RMC-RI), the Extreme norm Adversarial Training (E-AT) \cite{croce2022adversarial}, and the MSD Defense \cite{maini2020adversarial}. The evaluation methods encompass basic PGD adversarial attacks and Auto-Attack (AA) \cite{croce2020reliable} under $\ell_\infty$, $\ell_2$, $\ell_1$ norm perturbations and the MSD attack. The evaluation metrics include: (1) Standard accuracy on clean test data; (2) Robust accuracies under $\ell_\infty$, $\ell_2$, $\ell_1$-PGD adversarial attacks, MSD attack, and $\ell_\infty/\ell_2/\ell_1$ AA; (3) Accuracy on worst-case sample-wise (Union) using all three basic PGD adversarial attacks; and (4) DLR on $\ell_\infty$, $\ell_2$, $\ell_1$-PGD adversarial attacks for three types of perturbations and DLR on $\ell_\infty$, $\ell_2$ for two types of perturbations. All the following experiments are conducted on two NVIDIA RTX A100 GPUs.

\begin{figure*}[h]
  \centering
\subfloat[CIFAR100]{\includegraphics[trim=0 0 0 0,clip,width=0.25\textwidth]{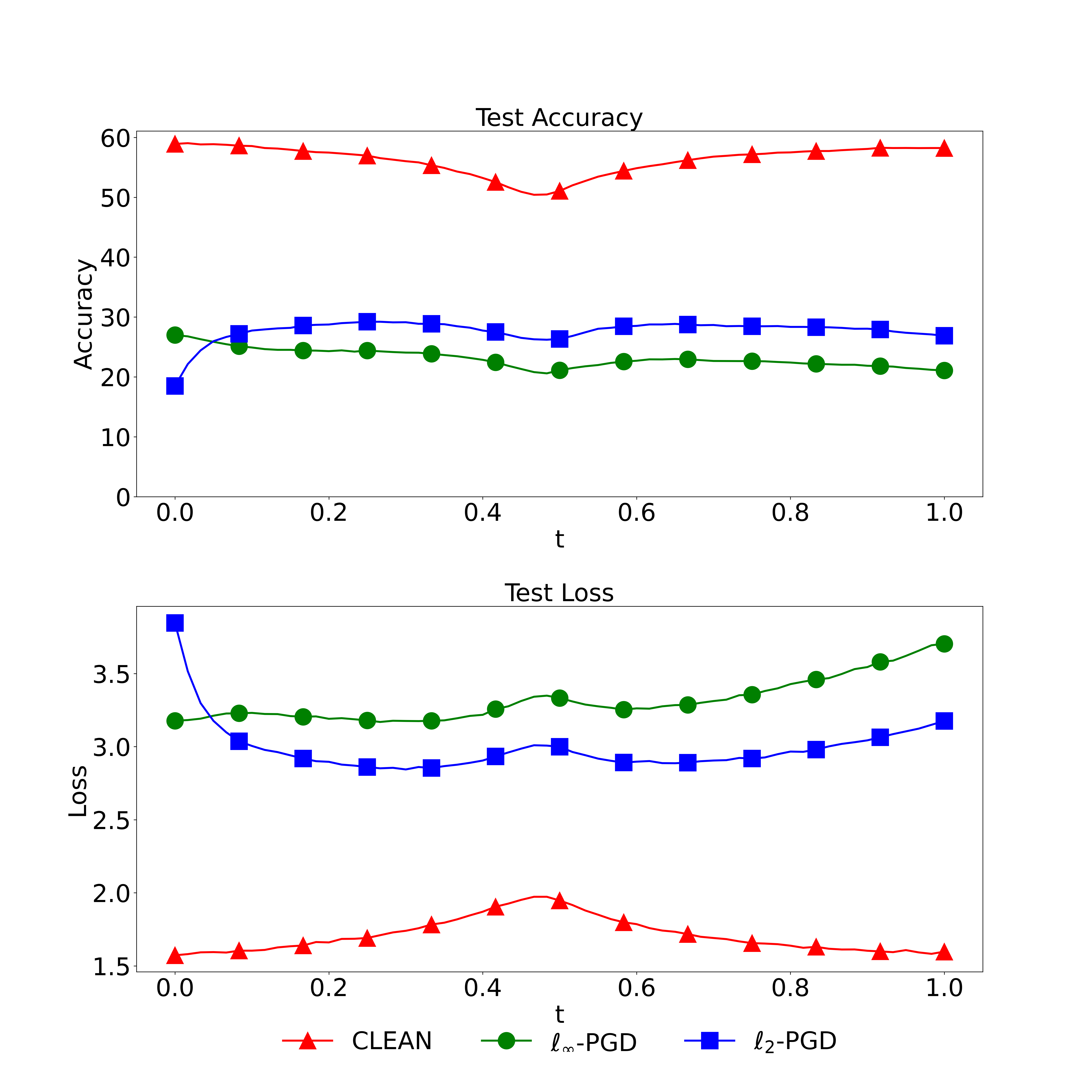}}
\subfloat[ImageNet-100]{\includegraphics[trim=0 0 0 0,clip,width=0.25\textwidth]{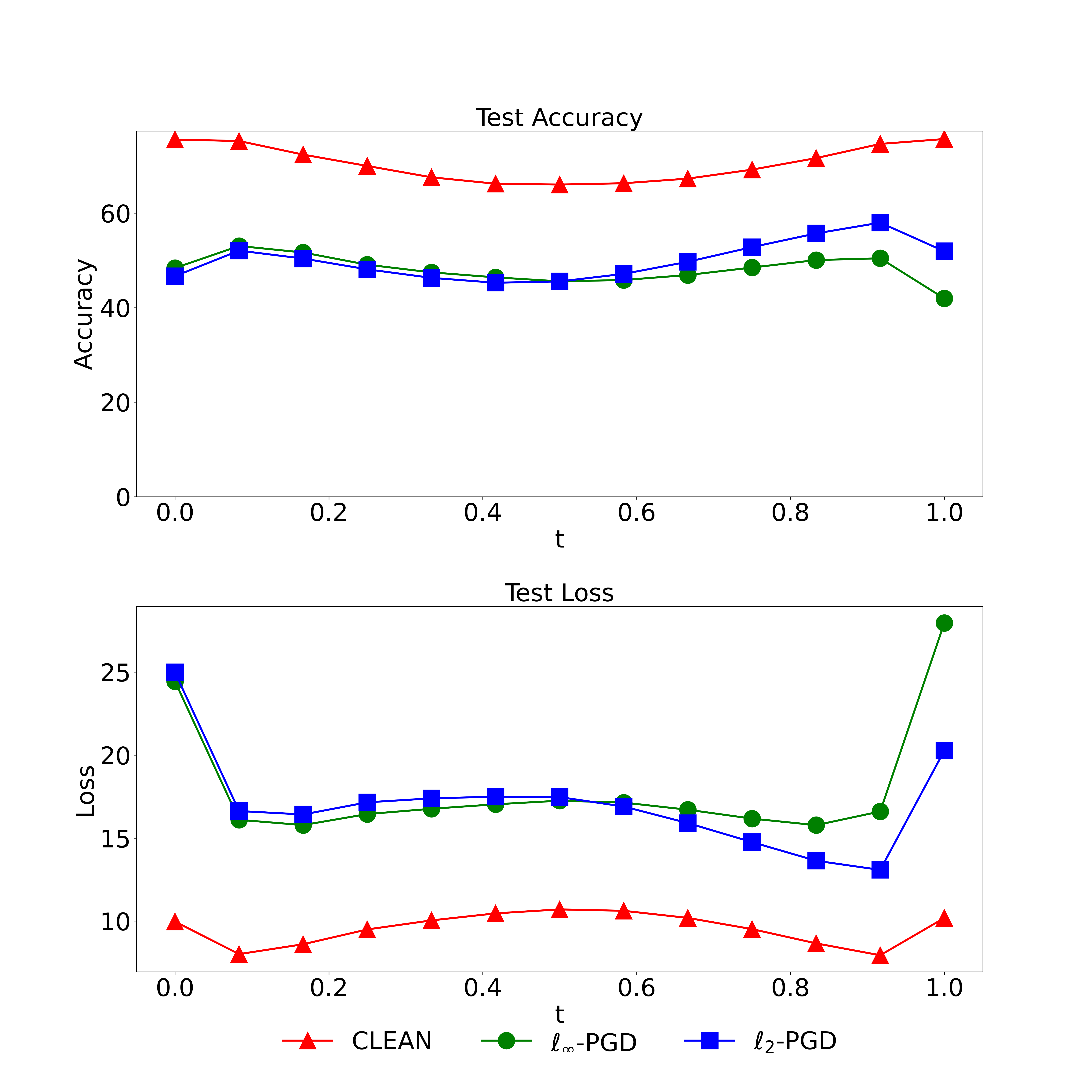}}
\subfloat[WideResNet-28-10]{\includegraphics[trim=0 0 0 0,clip,width=0.25\textwidth]{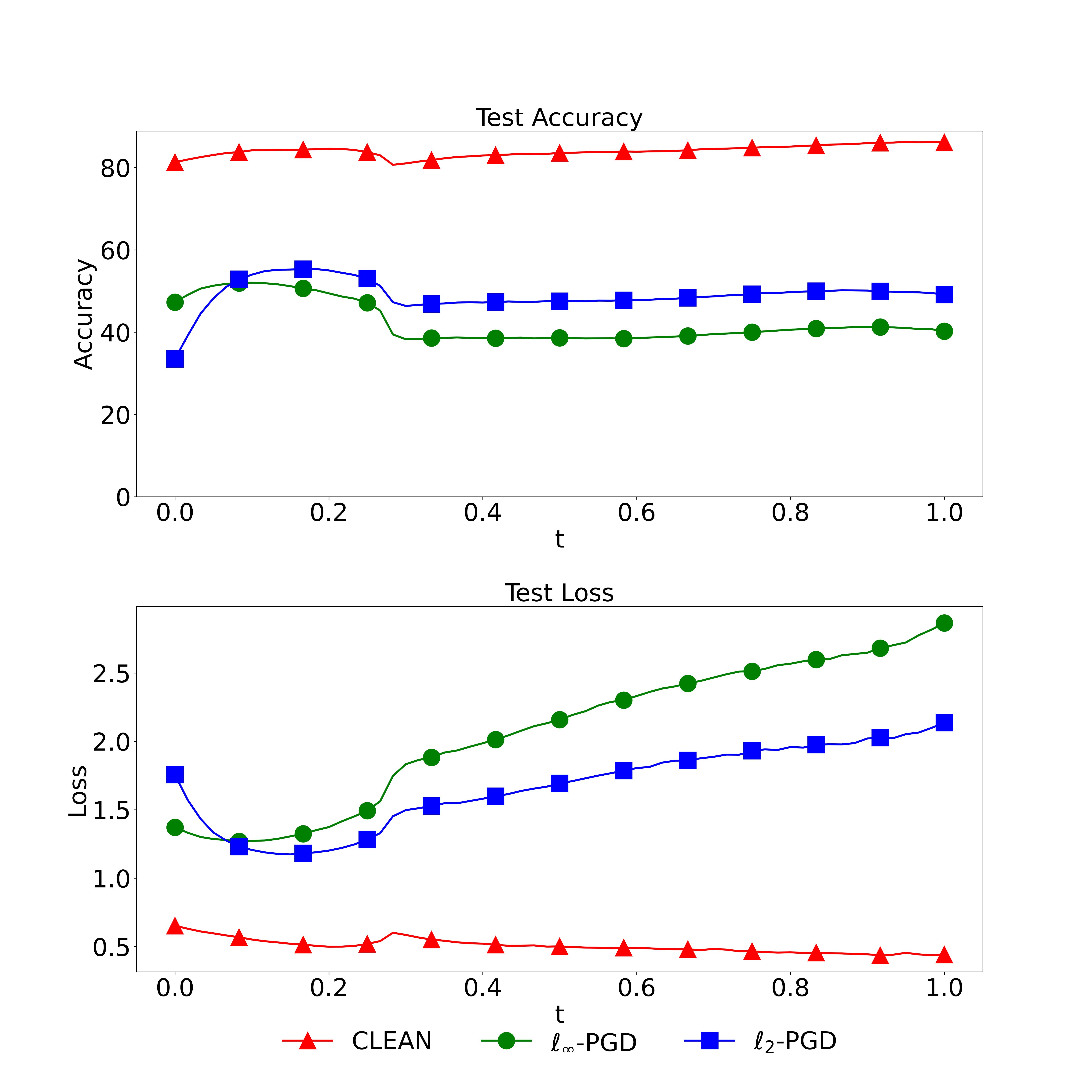}}
\subfloat[Vision Transformer-base]{\includegraphics[trim=0 0 0 0,clip,width=0.25\textwidth]{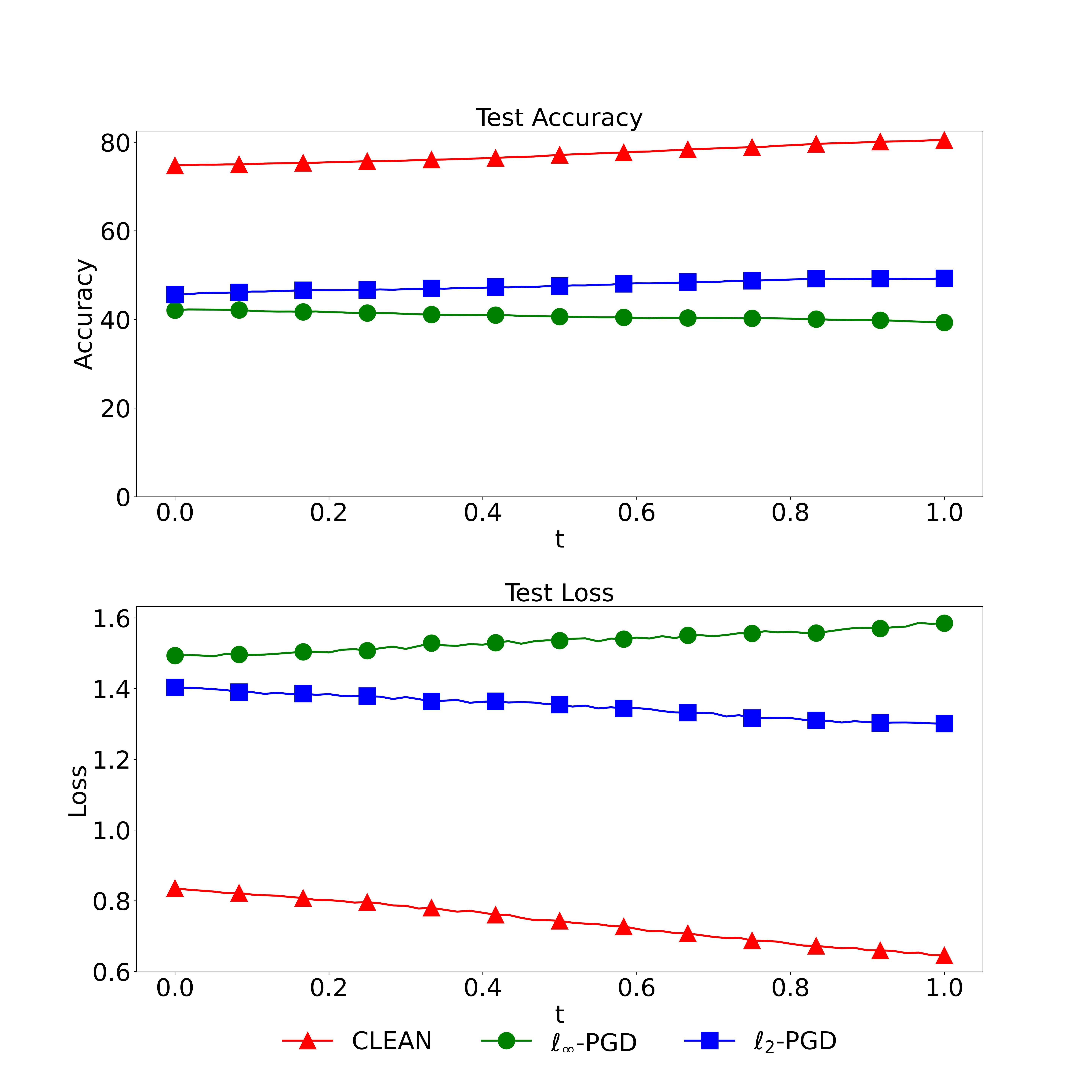}}
  \caption{{RMC is capable of identifying paths with points that have high DLR across various datasets and model architectures}. Figures (a) and (b), as well as (c) and (d), demonstrate that RMC performs effectively on the CIFAR-100 and ImageNet-100 datasets, as well as the WideResNet-28-10 and Vision Transformer-base architectures.}  
  \label{fig: rmc_vgg16_cifar100}
\end{figure*}

\subsection{A More Comprehensive Study of the Robust Mode Connectivity}

In this subsection, we aim to study the effectiveness of the proposed method \eqref{eq: mc_adv} on different models, architectures, and datasets. We will consider models trained under various settings. By default, we train endpoints' models 50/150 epochs and the paths are obtained by training an additional 50 epochs.

Here we evaluate the effectiveness of RMC on the CIFAR-100 and ImageNet-100 datasets, as well as the WideResNet-28-10 and Vision Transformer-base model architectures. We consider two types of perturbations that are generated from $\ell_\infty$ and $\ell_2$-PGD attacks. It can be seen from Fig.~\ref{fig: rmc_vgg16_cifar100} (a) and (b) that paths with high DLR points are obtained when CIFAR-10 is replaced with CIFAR-100 and ImageNet-100. Similarly, Fig.~\ref{fig: rmc_vgg16_cifar100} (c) and (d) demonstrate that paths with high DLR points are obtained when PreResNet110 is replaced with WideResNet-28-10 and Vision Transformer-base. These results underscore that RMC is versatile and can be effectively applied to various datasets and architectures.

\subsection{RMC with SRMC Modules}

\begin{figure}[h]
  \centering
  \includegraphics[trim=0 0 0 0,clip,width=.34\textwidth]{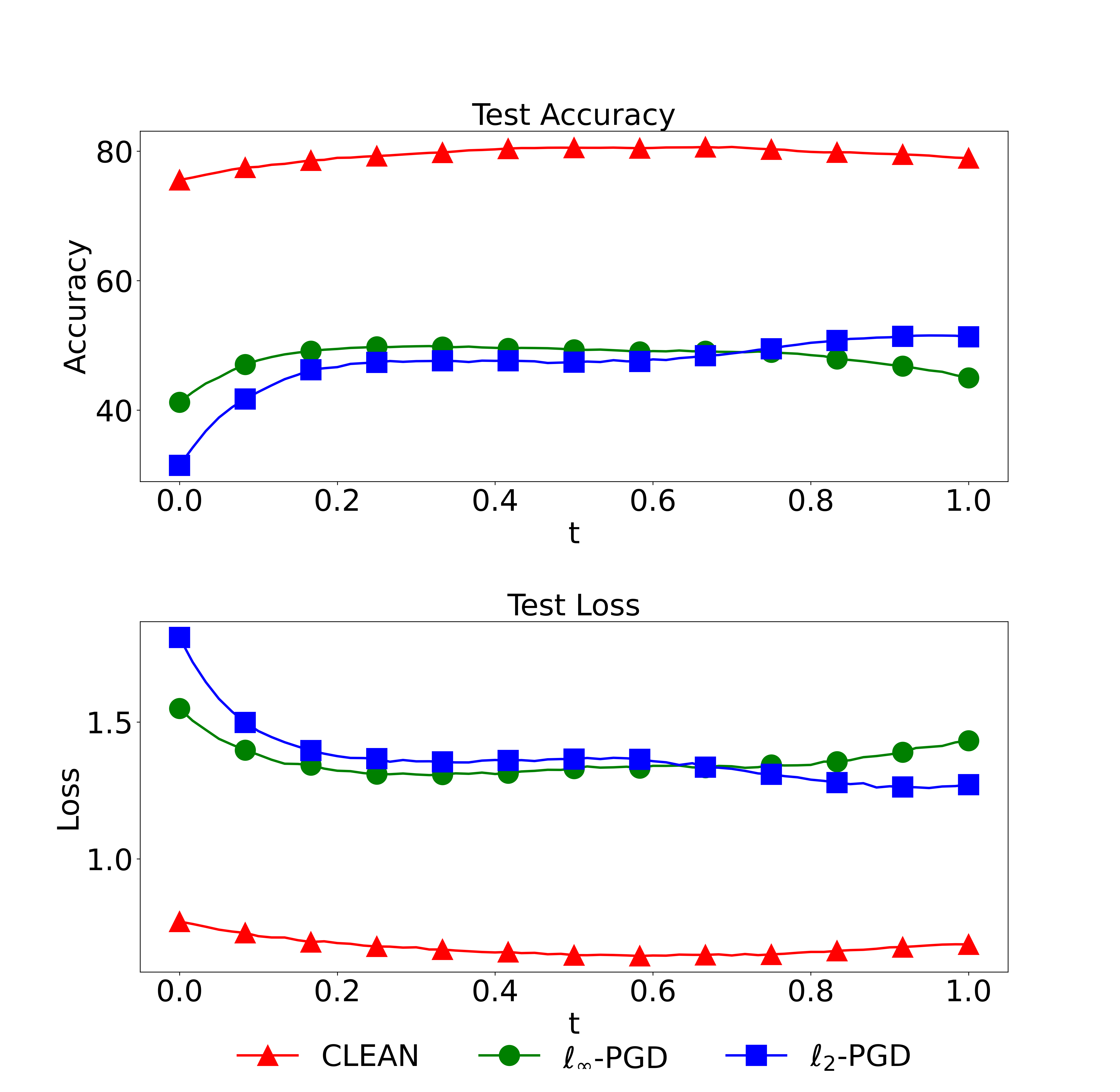}
  \caption{{A single SRMC module can also find paths with high DRL by connecting a $\ell_\infty$ model and a $\ell_2$ model}. } 
  \label{fig: adv_self_rmc}
\end{figure}

We then tested the proposed SRMC modules to expedite the RMC. Starting with a $\ell_\infty$-AT model, we trained an additional $\ell_2$-AT model and a $\ell_1$-AT model over 5 epochs. Subsequently, we connected each of these child models with the original $\ell_\infty$-AT model. The results, depicted in Fig.~\ref{fig: adv_self_rmc}, demonstrate the presence of paths with regions of high robustness under both connections. This indicates that we don't need to train all models from scratch to obtain the desired paths. For our Phase I experiments using PreResNet110 on CIFAR-10 with a single GPU, the process of learning RMC, which involved training two endpoint AT models for 150 epochs and the parameter curve for 50 epochs, took an average of $2750$ minutes. Learning SRMC under identical settings took $1780$ minutes. In the Vision Transformer setup with one GPU, learning RMC averaged $5785$ minutes, whereas learning SRMC in the same conditions required $3592$ minutes.

\begin{figure}[h]
  \centering
  \includegraphics[trim=0 0 0 0,clip,width=.46\textwidth]{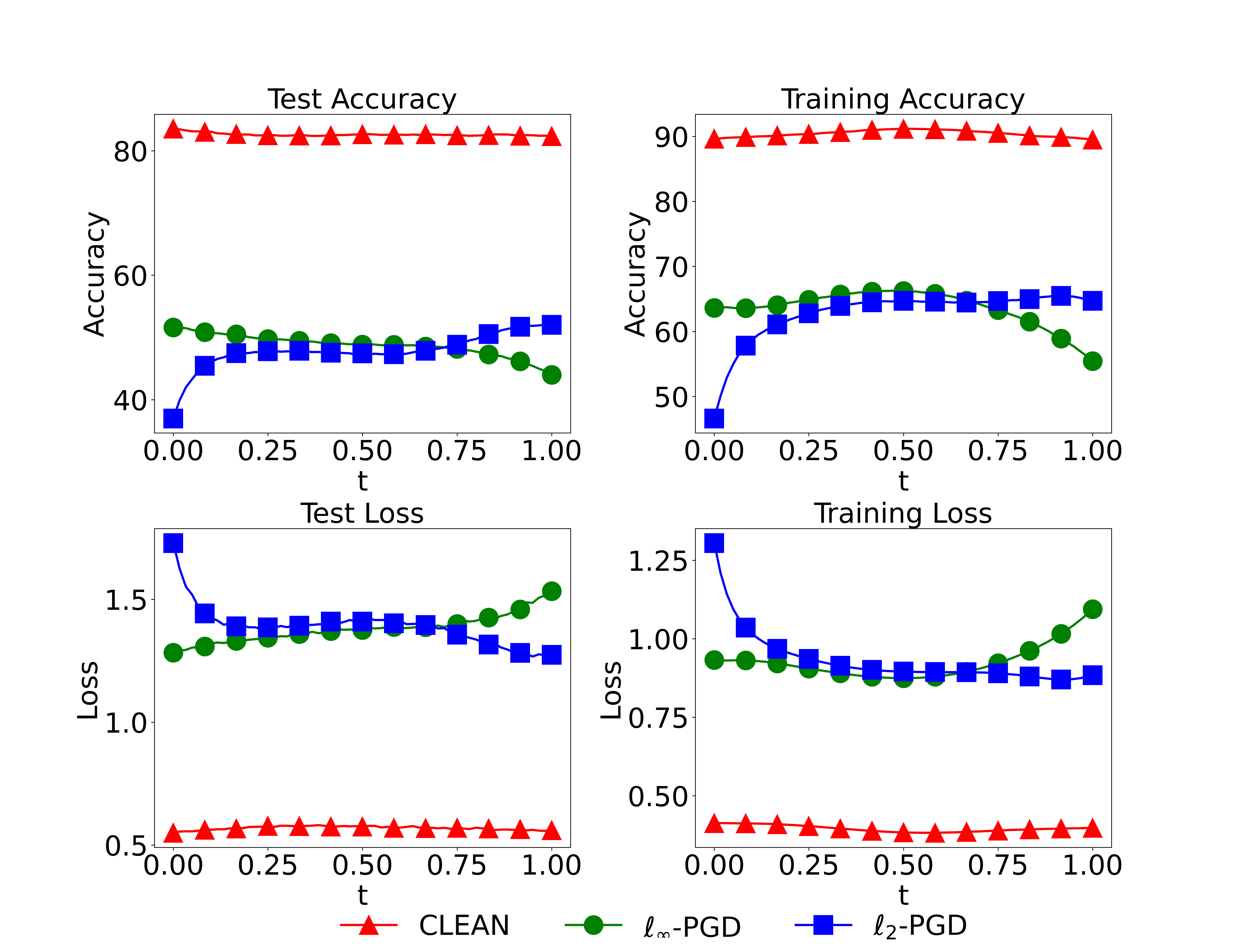}
  \caption{{RMC-based optimization considering two types of perturbations (one single mid-optimal point) can result in paths with smoother and higher DLR than the path in Fig.~\ref{fig: adv_mc} left panel}. The left (right) endpoint is an $\ell_\infty$-AT ($\ell_2$-AT) trained model starting from a single optimal point of a path connected between two models, which are trained by $\ell_\infty$-AT and $\ell_2$-AT for 50 epochs.} 
  \label{fig: adv_mc_opt1}
\end{figure}

\subsection{Robust Mode Connectivity-Based Optimization}\label{subsec: rmc_opt}

As introduced in Section~\ref{sec: opt}, Phase II is an enhanced optimization process based on units of RMC (Phase I). We show the effectiveness of RMC-based optimization (Phase II) below. Training epochs for all the experiments below are 200 (allow parallel computing).

\noindent\textbf{Optimization on two types of perturbations.} We first consider $\ell_\infty$ and $\ell_2$ norm perturbations. We train two models for 50 epochs under these two types of perturbations, then leverage RMC to find a path between the two models. Initializing from a single optimal point (randomly select from $t\in [0.77,0.83]$) on the curve, we train two models parallelly with $\ell_\infty$-AT and $\ell_2$-AT for 50 epochs. Finally, we plot the mode connectivity curve based on the two AT-trained models, as shown in Fig.~\ref{fig: adv_mc_opt1}. We obtain a smoother path with higher DLR than the path in Fig.~\ref{fig: adv_mc} left panel. The optimal point achieved is  $48.8\%$ at $t=0.72$.

\begin{figure}[h]
  \centering
  \includegraphics[trim=0 0 0 0,clip,width=.46\textwidth]{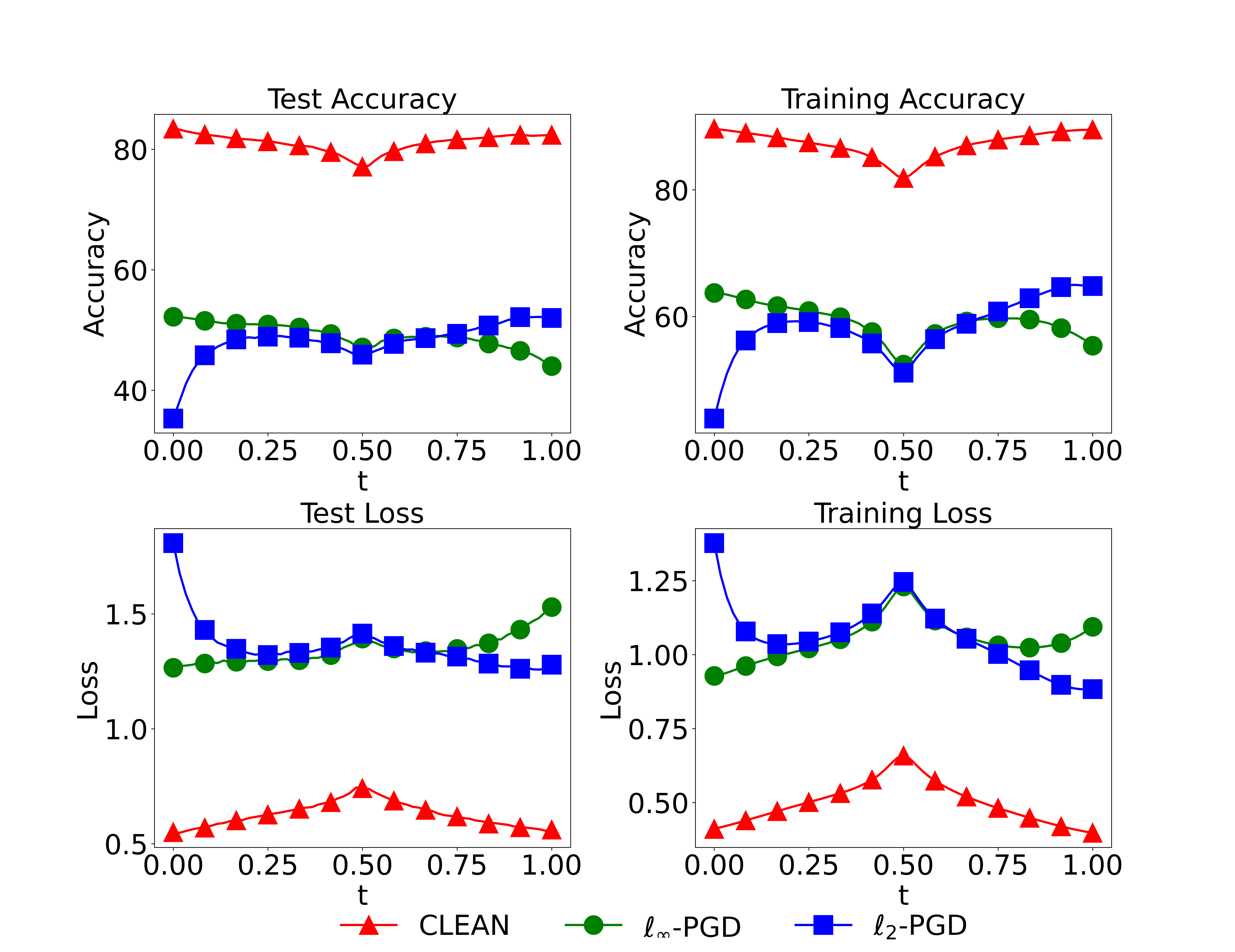}
  \caption{{RMC-based optimization considering two types of perturbations with two mid-optimal points is able to achieve higher robustness compared with only selecting a single mid-optimal point}. The left (right) endpoint is an $\ell_\infty$-AT ($\ell_2$-AT) trained model starting from two optimal points of a path connected between two models, which are trained by $\ell_\infty$-AT and $\ell_2$-AT for 50 epochs.} 
  \label{fig: adv_mc_opt12}
\end{figure}

Now instead of selecting a single optimal point, we randomly pick two optimal points in the ranges of $t\in [0.27,0.33]$ and $t\in [0.77,0.83]$, respectively. We train two models with $\ell_\infty$-AT and $\ell_2$-AT for 50 epochs starting from each initial point. We then plot the RMC curve based on the two AT-trained models, as shown in Fig.~\ref{fig: adv_mc_opt12}. The drop in accuracy observed at approximately $t=0.5$ is attributed to the small number of epochs (50) used in RMC. Increasing the number of epochs would result in smoother curves. The optimal point achieved in this optimization process is 48.89\% (DLR) at $t=0.71$, indicating that higher robustness can be improved by using a larger population with higher diversity.

\begin{figure}[h]
  \centering
  \includegraphics[trim=0 0 0 0,clip,width=.34\textwidth]{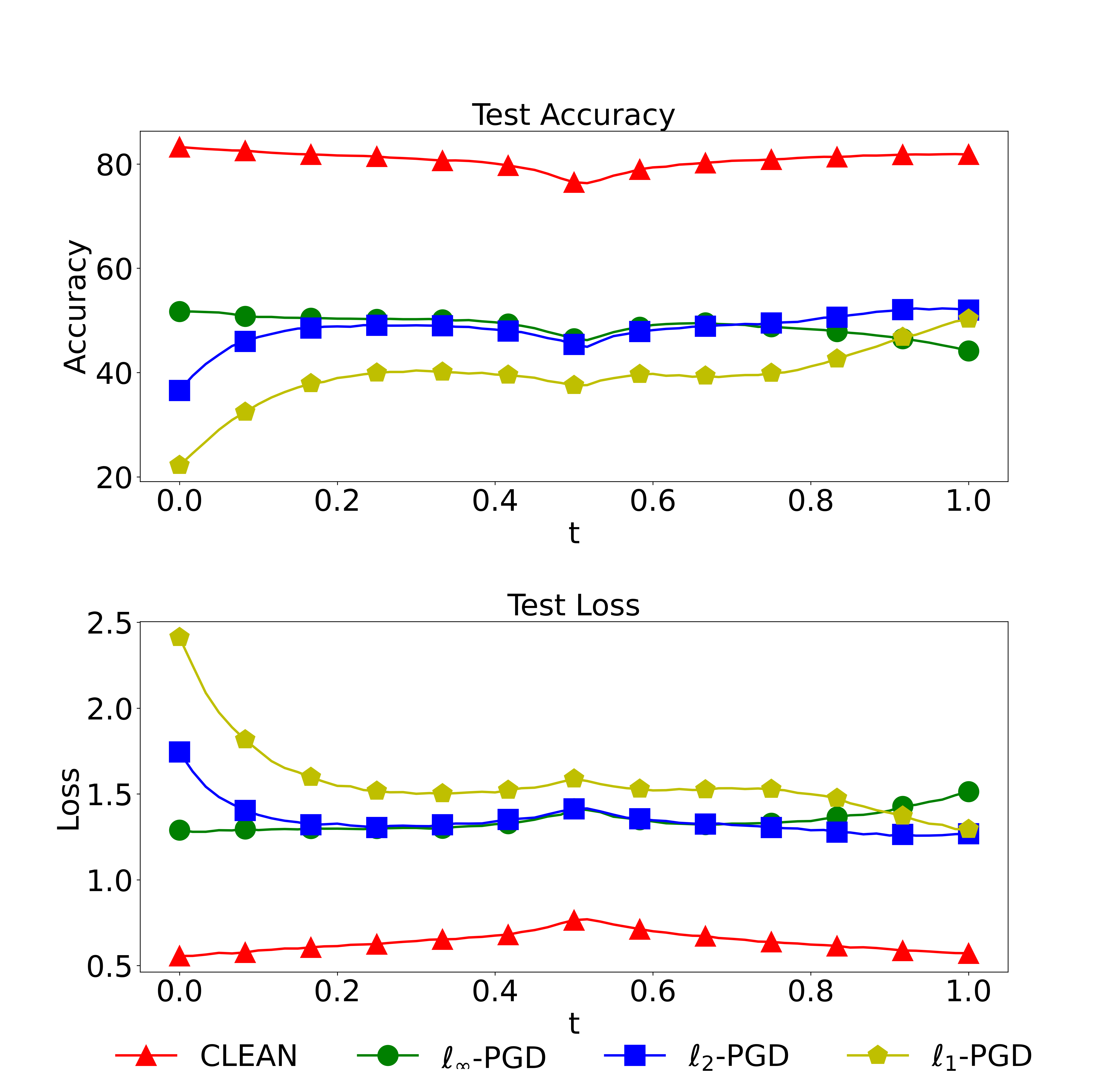}
  \caption{{RMC-based optimization considering three types of perturbations can further boost models' DLR against more types of attacks}. The two endpoints are trained by $\ell_\infty$-AT and $\ell_2$-AT for 50 epochs starting from the optimal points selected from two RMC curves.} 
  \label{fig: adv_mc_opt2}
\end{figure}

\noindent\textbf{Optimization on three types of perturbations.} We take one more step by considering the $\ell_1$ norm perturbation. The process is shown in Algorithm~\ref{alg: RMC_opt}. $T=50$ and we use 50 additional epochs to learn RMC. The results of the final connection are shown in Fig.~\ref{fig: adv_mc_opt2}. The trend of the $\ell_1$-PGD curve is increasing from left to right and the trend of the $\ell_2$-PGD curve from $t=0.7$ to $t=1$ is decreasing. There exists an optimal point with DLR$=46.21\%$ at $t=0.93$. RMC-based optimization in the case of three types of perturbations can further boost models' DLR against $\ell_\infty$, $\ell_1$, $\ell_2$ adversarial attacks. Additionally, one can select multiple models from the curve and use ensemble methods to further improve performance. 

\begin{figure}[ht]

\begin{minipage}[b]{.48\linewidth}
  \centering
  \centerline{\includegraphics[width=5.0cm]{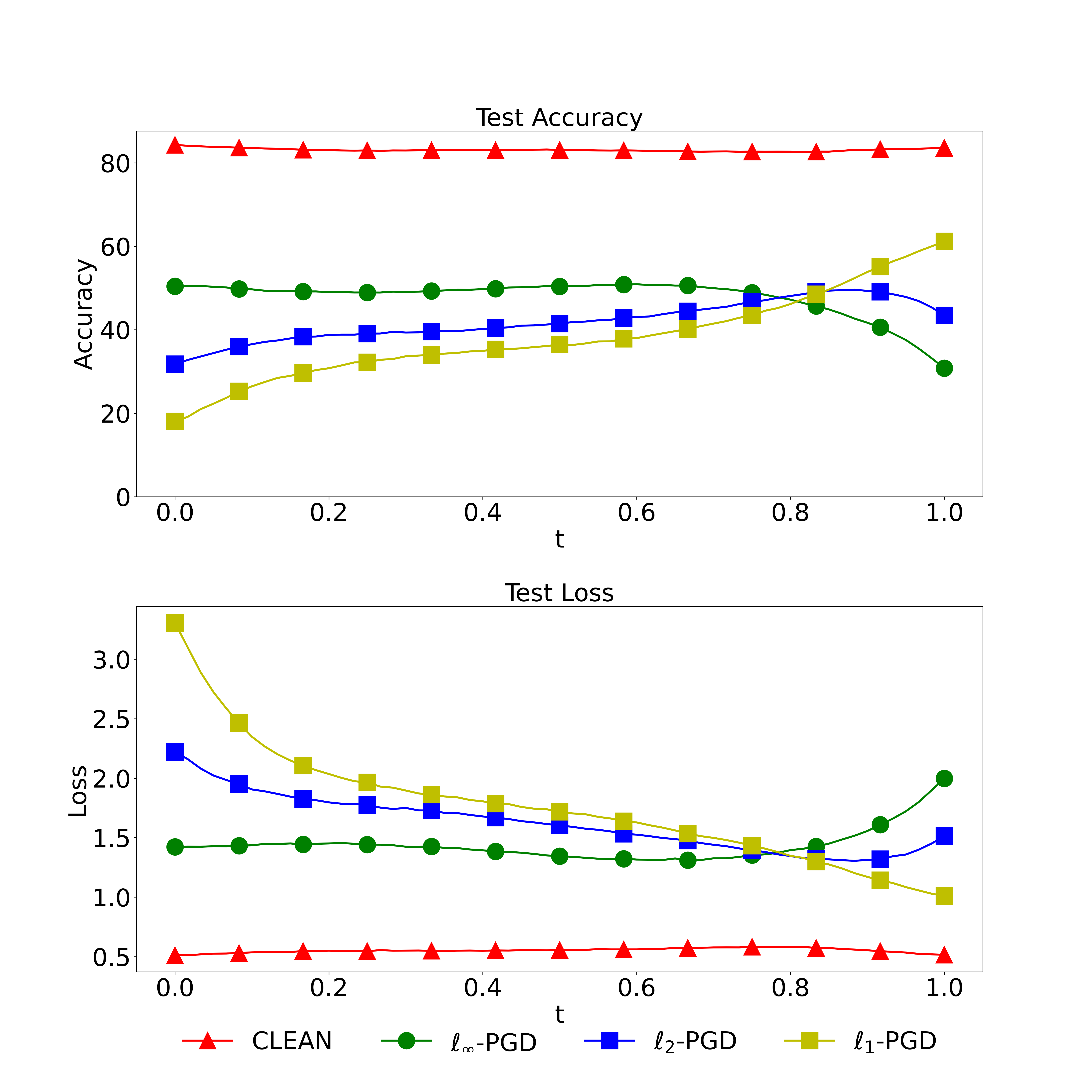}}
  \centerline{(a) CIFAR-10}\medskip
\end{minipage}
\begin{minipage}[b]{.48\linewidth}
  \centering
  \centerline{\includegraphics[width=5.0cm]{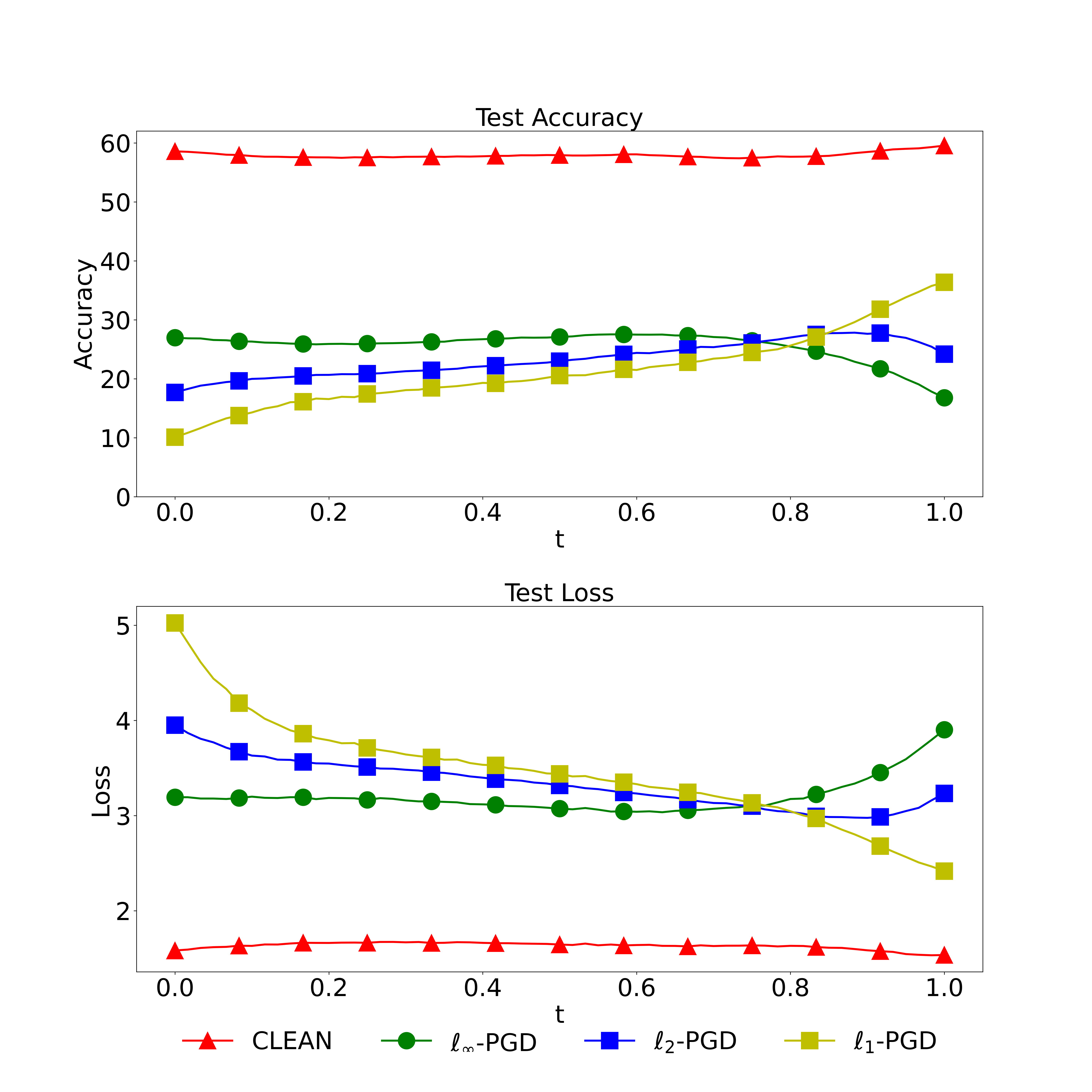}}
  \centerline{(b) CIFAR-100}\medskip
\end{minipage}
\begin{minipage}[b]{.48\linewidth}
  \centering
  \centerline{\includegraphics[width=5.0cm]{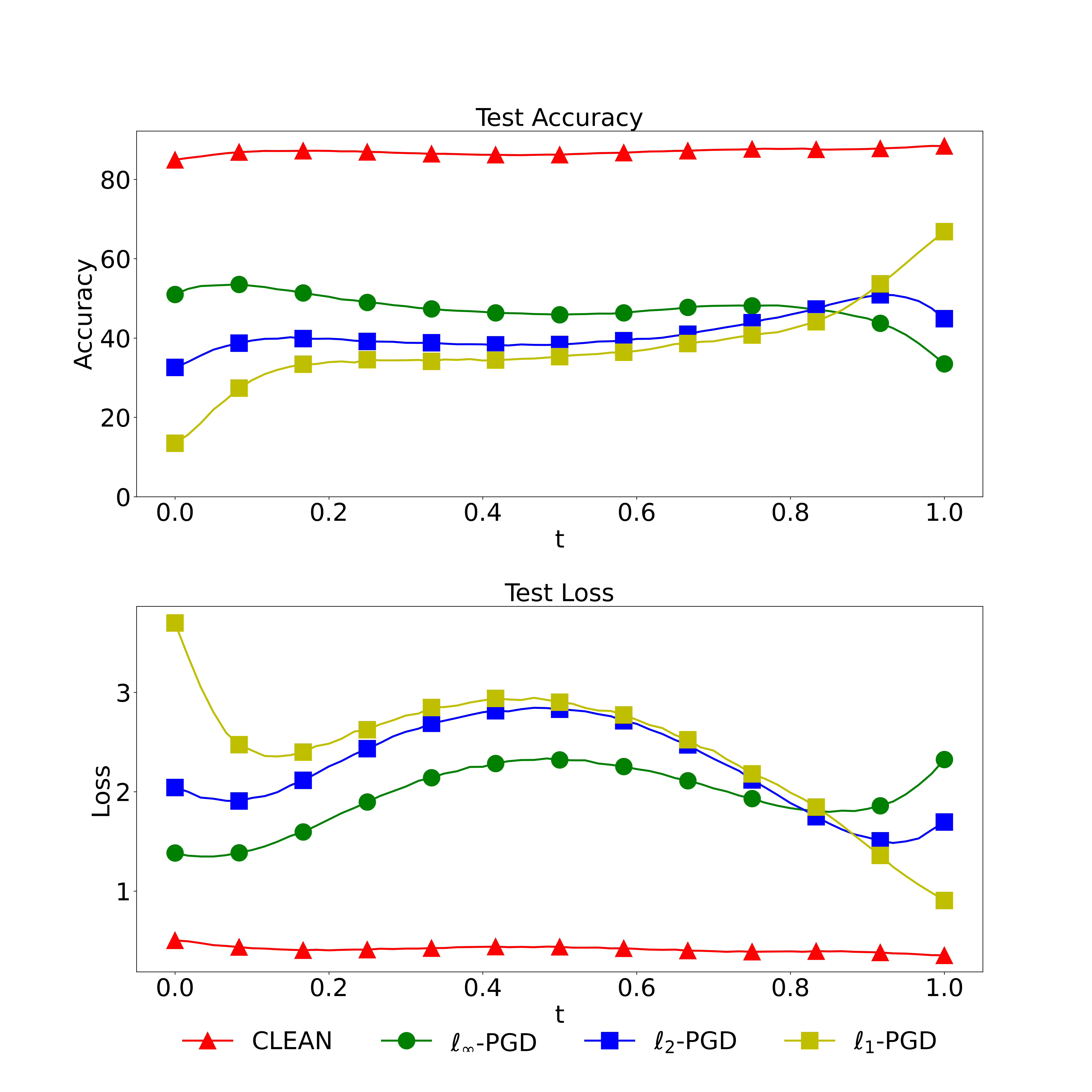}}
  \centerline{(c) WideResNet-28-10}\medskip
\end{minipage}
\hfill
\begin{minipage}[b]{0.48\linewidth}
  \centering
  \centerline{\includegraphics[width=5.0cm]{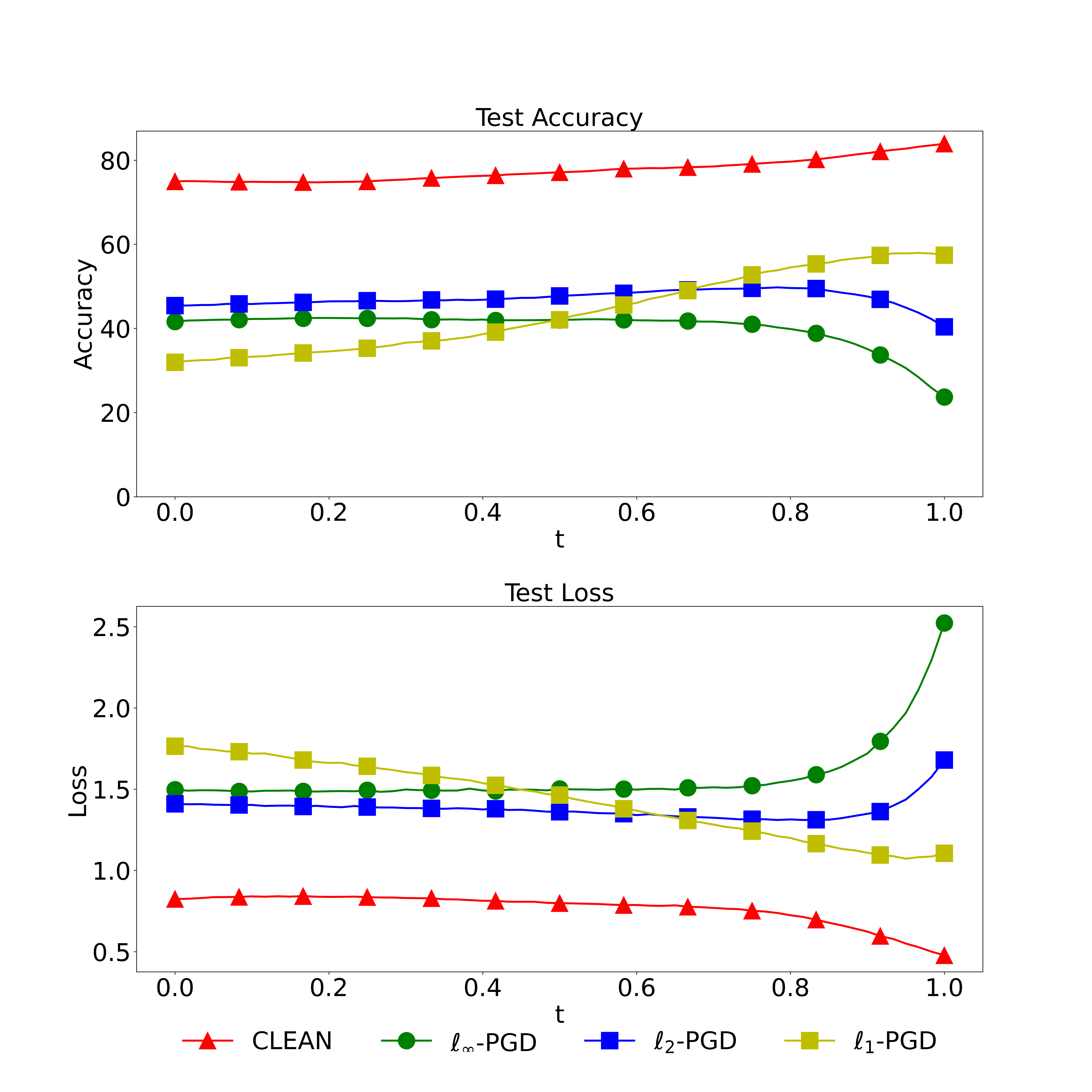}}
  \centerline{(d) ViT-base}\medskip
\end{minipage}
\caption{ERMC can find paths with high robustness against $\ell_\infty/\ell_2/\ell_1$ attacks by connecting a $\ell_\infty$ model and a $\ell_1$ model. The effectiveness of ERMC is validated on different datasets and model architectures.}
\label{fig: adv_self_rmc2}
\end{figure}

\subsection{Results on ERMC}

In ERMC, the models situated at the left and right endpoints undergo different training processes. The left endpoint model receives training with $\ell_\infty$-AT, whereas the right endpoint model is subsequently refined with AT fine-tuning focused on $\ell_1$-AT. These procedures' outcomes are illustrated in Fig.~\ref{fig: adv_self_rmc2}. The observations from this process are twofold: Firstly, ERMC demonstrates commendable performance across all tested datasets and architectures. Secondly, the process of fine-tuning has a noticeable impact on the models' inherent robustness. Models at each endpoint exhibit a high degree of robustness to the type of perturbations they were trained against (i.e., $\ell_\infty$ for the left endpoint and $\ell_1$ for the right) yet they show a relative vulnerability to the alternate type of perturbations (i.e., $\ell_1$ for the left endpoint and $\ell_\infty$ for the right).

\begin{table*}[!t]
\caption{Our Methods Achieve State-of-the-Art DLR Levels Under Various Perturbations. Furthermore, our methods consistently achieve the highest accuracy under Union, $\ell_\infty/\ell_2/\ell_1$ AA \cite{croce2020reliable} (with the lowest accuracies being denoted by braces), and MSD. For methods utilizing two types of perturbations, we compare them using DLR only under $\ell_\infty$-PGD and $\ell_2$-PGD attacks, marking the DLR (representing the lowest accuracy) with an underline. For those employing three types of perturbations, we assess them across all metrics, marking the DLR (the lowest accuracy) under the three basic $\ell_p$ attacks with an overline.
}
\label{tab: main}
\begin{center}

\resizebox{0.99\textwidth}{!}{
\begin{tabular}{l||c|c|c|c|c|c|c|c}
\hline
\hline
& Standard Accuracy & $\ell_\infty$-PGD ($\delta=8/255$) & $\ell_2$-PGD ($\delta=1$) & $\ell_1$-PGD ($\delta=12$) & DLR & Union & AA ($\ell_\infty/\ell_2/\ell_1$) \cite{croce2020reliable} & MSD \\
\hline
\begin{tabular}[c]{@{}c@{}} $\ell_\infty$-AT \cite{madry2018towards}   \end{tabular}  &   85.00\% & 49.03\% & 29.66\%  & 16.61\% &  / & 21.85\%  & 46.02\%/20.86\%/\{10.45\%\} & 15.27\% \\
\hline
\begin{tabular}[c]{@{}c@{}} MSD - Defense \\ (\textbf{two} types of pert)  \end{tabular}  &  81.61\% & 48.57\% & \underline{45.92\%} & 35.64\% & 45.92\% & 34.37\% & 46.6\%/42.13\%/\{31.55\%\} & 45.72\%  \\
\hline
\begin{tabular}[c]{@{}c@{}} RMC-RI \\ (\textbf{two} types of pert)  \end{tabular}  &  63.08\% & \underline{37.4\%} & 38.44\% & 30.47\% & 30.47\% & 29.22\% & 36.85\%/37.17\%/\{28.33\%\} & 36.9\%  \\
\hline
\begin{tabular}[c]{@{}c@{}} RMC \\ (ours, \textbf{two} types of pert)  \end{tabular}  &  80.90\%  & \underline{48.19\%} & 48.63\% & 38.05\% & 48.19\%  & 36.3\% & 46.74\%/45.16\%/\{34.4\%\} & 46.52\%  \\
\hline
\begin{tabular}[c]{@{}c@{}} RMC-based optimization \\ (ours, \textbf{two} types of pert)  \end{tabular}  &  81.36\%  & \underline{48.89\%} & 49.03\% & 38.83\% & 48.89\%  & 36.86\% & 47.66\%/45.73\%/\{35.18\%\} & 47.18\%  \\
\hline
\begin{tabular}[c]{@{}c@{}} MSD - Defense \cite{maini2020adversarial} \\ (\textbf{three} types of pert)  \end{tabular}  &  81.35\% & $\overline{40.14\%}$ & 48.58\% & 47.50\% & 40.14\% & 38.35\% & \{37.87\%\}/45.9\%/45.27\% & 38.20\%  \\
\hline
\begin{tabular}[c]{@{}c@{}} E-AT  \cite{croce2022adversarial} \\ (\textbf{three} types of pert)  \end{tabular}  &  79.3\% & $\overline{44.07\%}$ & 49.12\% & 49.82\% & 44.07\% & 41.08\% & \{41.41\%\}/46.5\%/47.82\% & 42.67\%  \\
\hline
\begin{tabular}[c]{@{}c@{}} RMC-based optimization \\ (ours, \textbf{three} types of pert)  \end{tabular}  &  81.76\% & $\overline{46.21\%}$ & 51.86\% & 46.23\% & 46.21\% & 41.47\% & 44.58\%/49.35\%/\{43.42\%\} & 44.75\%   \\
\hline
\begin{tabular}[c]{@{}c@{}} RMC-based optimization \\ 5-model ensemble \\ (ours, \textbf{three} types of pert)  \end{tabular}  &  78.35\% & 55.91\% & 56.78\% & $\overline{51.05\%}$ & 51.05\% & 49.39\% & 50.15\%/49.85\%/{\{\bf{49.83\%}\}} & 48.79\%   \\
\hline
\begin{tabular}[c]{@{}c@{}} RMC-based optimization \\ with SRMC modules \\ (ours, \textbf{three} types of pert)  \end{tabular}  &  80.39\%  & $\overline{46.10\%}$ & 48.92\% & 46.39\% & 46.10\%  & 42.03\% & 44.95\%/46.66\%/\{43.91\%\} & 45.07\%   \\
\hline
\begin{tabular}[c]{@{}c@{}} ERMC-1 \\ (ours, \textbf{three} types of pert) \end{tabular}  &  82.66\%  &   $\overline{46.54\%}$ & 48.76\% & 47.06\% & 46.54\% &  41.94\% & 44.88\%/45.88\%/\{43.97\%\} & 44.88\%   \\
\hline
\begin{tabular}[c]{@{}c@{}} ERMC-3 \\ (ours, \textbf{three} types of pert)  \end{tabular}  &  79.61\%  & 49.29\% & 51.32\% & $\overline{48.49\%}$ & 48.49\% &  45.27\% & \{42.88\%\}/44.57\%/47.37\% & 43.31\%   \\
\hline
\begin{tabular}[c]{@{}c@{}} ERMC-5 \\ (ours, \textbf{three} types of pert) \end{tabular}  &  79.41\%  & 55.46\% & 57.28\% & $\overline{53.97\%}$ & \bf{53.97\%} &  \bf{51.41\%} & {\{49.33\%}\}/50.55\%/52.41\% & {\bf{49.78\%}}   \\
\hline
\hline
\end{tabular}}
\end{center}
\end{table*}

\subsection{A Comprehensive Comparison}

For MSD Defense, RMC-RI, RMC, and RMC-based optimization (when considering only two types of perturbations), we evaluated them using the $\ell_\infty$-PGD and $\ell_2$-PGD attacks, given that the $\ell_1$-PGD attack was not considered during training. The DLR (representing the lowest robust accuracy) for these attacks is indicated with an underline. 

For methods that employ three types of perturbations, we assessed them under the $\ell_\infty$-PGD, $\ell_2$-PGD, $\ell_1$-PGD, AA, MSD attacks, and the union metric. In RMC-based optimization, a 5-model ensemble means that we select five models from the curve shown in Fig.~\ref{fig: adv_mc_opt2} and ensemble them. For the model ensemble, the model selection thresholds are set at $\alpha_\infty=37\%$ for $\ell_\infty$ robustness and $\alpha_1=43\%$ for $\ell_1$ robustness. We determine the lowest accuracy among the $\ell_\infty$, $\ell_2$, and $\ell_1$ norms within the AA framework and mark it with braces. The DLRs (lowest robust accuracies) for the $\ell_\infty$-PGD, $\ell_2$-PGD, and $\ell_1$-PGD attacks are denoted with an overline. Additionally, we emphasized the highest accuracy values in the union, AA, and MSD columns.

From Table~\ref{tab: main}, the following observations can be made: (1) RMC with two types of perturbations outperforms MSD with two types of perturbations on DLR by $2.27\%$ and also surpasses RMC-RI by $10.79\%$.; (2) The RMC-based optimization with two types of perturbations yields a slightly higher DLR than RMC and excels over RMC in all other metrics; (3) When considering three types of perturbations, RMC-based optimization surpasses both MSD Defense and E-AT in DLR by $6.07\%$ and $2.14\%$. Moreover, it exhibits accuracy improvements of $3.12\%$, $5.55\%$, and $6.55\%$ ($0.39\%$, $2.01\%$, and $2.08\%$) over MSD Defense (and E-AT) under the Union, AA, and MSD Attack metrics, respectively; (4) The RMC-based optimization method shows a trade-off between clean accuracy and DLR. However, its clean accuracy drop of $3.24\%$, when benchmarked against $\ell_\infty$-AT, is less severe than that observed in other defense methods like MSD Defense and E-AT. Currently, the model selection process in RMC prioritizes robustness (DLR) over clean accuracy, but this could be adjusted in future implementations to achieve a better balance between the two; (5) The RMC-based optimization with SRMC modules can achieve similar (slightly lower) DLR performance compared to the RMC-based optimization with three types of perturbations, and even has slightly higher accuracy under the AA and Union metric; (6) Using a multi-model ensemble method can further enhance the performance of RMC-based optimization; (7) ERMC-1 reaches similar performance as RMC-based optimizataion. As the number of models $n$ increases, the performance of ERMC correspondingly improves. When $n$ reaches 5, ERMC-5 outperforms all other methods in terms of accuracy improvements under DLR, Union, and MSD.

In conclusion, our proposed Robust Mode Connectivity-Oriented Adversarial Defense shows remarkable performance across a variety of metrics. The RMC-based optimization (Phase II) delivers a higher DLR compared to RMC (Phase I) alone. ERMC can achieve high robustness by only conducting one RMC process. On the whole, the Robust Mode Connectivity-Oriented Adversarial Defense introduces a novel defense paradigm rooted in population-based optimization, effectively enhancing the robustness of Neural Networks (NNs).

\section{Conclusion}\label{sec: conclusion}
In this work, we introduced a Robust Mode Connectivity (RMC)-oriented adversarial defense framework that leverages population-based optimization to strengthen neural networks against diversified $\ell_p$ attacks. Our two-phase design enables the discovery of robust paths (Phase I) and systematic selection of high-performing models through RMC-based optimization (Phase II). To improve efficiency, we further proposed the Efficient Robust Mode Connectivity (ERMC), which combine theoretical guarantees with practical scalability. Extensive experiments across CIFAR-10, CIFAR-100, and ImageNet-100, as well as multiple architectures, demonstrated that our approach consistently outperforms existing methods, achieving superior diversified $\ell_p$ robustness while maintaining competitive accuracy. Overall, this work establishes population-based mode connectivity as a powerful and generalizable principle for adversarial defense. Future directions include extending RMC to large-scale foundation models, integrating it with certified robustness techniques, and exploring applications in safety-critical domains such as healthcare and power systems.

\bibliographystyle{ieeetr}
\bibliography{ref_adv,ref_new}

\newpage
\twocolumn
\pagestyle{empty}

\vfill
\end{document}